\RequirePackage{fix-cm}
\documentclass[twocolumn]{svjour3}     
\smartqed  
\usepackage{fix-cm}
\usepackage{graphicx}
\usepackage{latexsym}
\usepackage{ifpdf}
\usepackage{booktabs}
\usepackage{mathrsfs}
\usepackage{amssymb}
\newcommand{\etal}{\textit{et al.}} 
\newcommand{\ie}{\textit{i.e. }}
\newcommand{\eg}{\textit{e.g. }}
\usepackage{epsfig}
\usepackage{multirow}
\usepackage{cite}
\usepackage{graphicx}
\usepackage{amsmath}
\usepackage{mathtools}
\usepackage{algorithmic}
\usepackage{array}
\usepackage[caption=false,font=footnotesize]{subfig}
\usepackage{url}
\usepackage{xcolor}
\usepackage{caption}

\begin{document}

\title{\textcolor{black}{EAML: Ensemble Self-Attention-based Mutual Learning Network for Document Image Classification}
}

\titlerunning{EAML}

\author{Souhail Bakkali         \and
        Zuheng Ming             \and 
        Mickaël Coustaty        \and 
        Marçal Rusiñol        
}

\authorrunning{Souhail Bakkali et al.}

\institute{Souhail Bakkali \at
           L3i, University of La Rochelle, La Rochelle, France \\
           \email{souhail.bakkali@univ-lr.fr}          
           \and
           Zuheng Ming \at
           L3i, University of La Rochelle, La Rochelle, France \\
           \email{zuheng.ming@univ-lr.fr}          
           \and
           Mickaël Coustaty \at
           L3i, University of La Rochelle, La Rochelle, France \\
           \email{mickael.coustaty@univ-lr.fr}           
           \and
           Marçal Rusiñol \at
           AllRead Machine Learning Technologies, Barcelona, Spain\\
           \email{marcal@allread.ai}          
           \and
}

\date{Received: 18/11/2020 / Accepted: 29/03/2021}

\maketitle

\begin{abstract}
In the recent past, complex deep neural networks have received huge interest in various document understanding tasks such as document image classification and document retrieval. As many document types have a distinct visual style, learning only visual features with deep CNNs to classify document images have encountered the problem of low inter-class discrimination, and high intra-class structural variations between its categories. In parallel, text-level understanding jointly learned with the corresponding visual properties within a given document image has considerably improved the classification performance in terms of accuracy.
In this paper, we design a self-attention-based fusion module that serves as a block in our ensemble trainable network. It allows to simultaneously learn the discriminant features of image and text modalities throughout the training stage. \textcolor{black}{Besides, we encourage mutual learning by transferring the positive knowledge between image and text modalities during the training stage. This constraint is realized by adding a truncated-Kullback–Leibler divergence loss (Tr-KLD$_{Reg}$) as a new regularization term, to the conventional supervised setting.}
To the best of our knowledge, this is the first time to leverage a mutual learning approach along with a self-attention-based fusion module to perform document image classification. The experimental results illustrate the effectiveness of our approach in terms of accuracy for the single-modal and multi-modal modalities. Thus, the proposed ensemble self-attention-based mutual learning model outperforms the state-of-the-art classification results based on the benchmark RVL-CDIP and Tobacco-3482 datasets.


\end{abstract}

\begin{figure*}[ht]
\centering
  \centerline{\includegraphics[width=\linewidth]{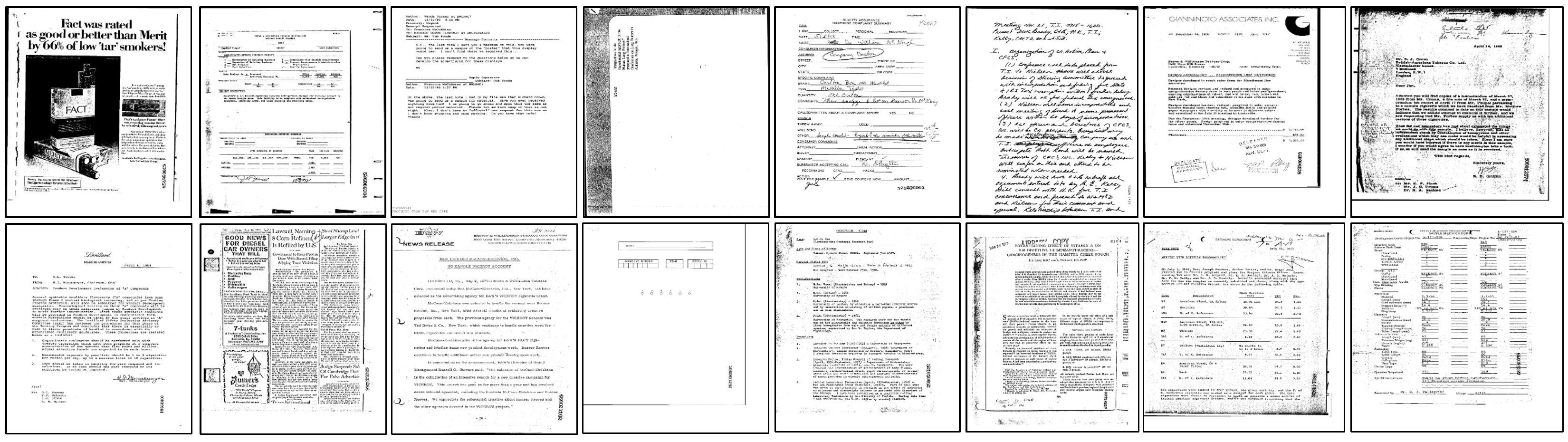}}
    \caption{Samples of different document classes in the RVL-CDIP dataset. From left to right: \textit{Advertisement, Budget, Email, File folder, Form, Handwritten, Invoice, Letter, Memo, News article, Presentation, Questionnaire, Resume, Scientific publication, Scientific report, Specification}}
    \label{fig:dataset}
\smallskip
\end{figure*}

\section{Introduction}
\label{sec:intro}
Deep Learning has provided compelling results in various document understanding problems such as document  retrieval, information extraction, and document image classification. Thanks to its impressive performance over a large number of tasks, this area has been explored extensively. Existing works covered several techniques including document binarization~\cite{Afzal2015DocumentIB, PastorPellicer2015InsightsOT}, layout analysis~\cite{PastorPellicer2016CompleteSF, Seuret2017PCAInitializedDN}, and structural similarity constraints~\cite{Chen2006ASO} for many document analysis tasks. However, to ensure a good generalization, many deep neural networks with large amount of parameters have been used for document image classification in order to extract the most relevant visual features~\cite{LeCun1998GradientbasedLA}. 

Unlike the general images from the ImageNet dataset ~\cite{Russakovsky2015ImageNetLS}, document images have a distinct visual style. Therefore, numerous studies on document processing tasks have used transfer learning. It has shown to be effective on boosting the classification performance of document images~\cite{Harley2015EvaluationOD, Afzal2015DeepdocclassifierDC, 8545630}, whereas randomly initialized networks are under-performing~\cite{Kang2014ConvolutionalNN}. Additionally, from the perspective of a natural language processing classifier, document images can be categorized into various classes based on their textual content processed by an Optical Character Recognition (OCR) system~\cite{Qian2010ANA, UlHasan2015ASL}. Yang \etal~\cite{Yang2017LearningTE} presented a neural network to extract semantic enriched information from textual content based on a word embedding mechanism. Also, Appiani \etal ~\cite{Appiani2001AutomaticDC} described a system that exploits a structural analysis approach to characterize and automatically index heterogeneous documents with variable layout, by determining the class of the document image based on reliable automatic information extraction methods.

\textcolor{black}{Nevertheless, the challenge of document images remains in their wide range of visual variability, where documents from the same category might have different spatial properties.}
Due to their particular visual style, relying on deep convolutional networks to extract visual properties to perform document image classification might fail to distinguish between highly correlated classes. The intra-class variability of document images might be even larger than the inter-class variability, where two or multiple document images of different categories can be visually, and in terms of their textual content, closer than two or multiple documents from the same category. This level of intra-class variability can be mitigated by introducing the latent semantic information from the text corpus within the document image. Once the visual features of the image modality and the textual features of the text modality are extracted, they are leveraged into a multi-modal network to combine both feature vectors into one feature vector based on a feature fusion methodology~\cite{Dauphinee2019ModularMA, audebert2019multimodal, Noce2016EmbeddedTC}.
Typically, multi-modal methods for document image classification rely on image and text modalities. They contain two or an ensemble of deep networks which are pre-trained on large-scale datasets to extract discriminate features from the input data. With such approaches, the learning process of the image modality and the text modality is still independent one from another. The output features of both modalities are subsequently combined together to perform an ensemble trainable document image classification network~\cite{Ferrando2020ImprovingAA, Xu2020LayoutLMPO, Asim2019TwoSD, Xu2020LayoutLMv2MP}. Yet, these independent learning approaches might be enhanced if the visual and the textual features share some mutual information between them.

In this paper, we propose an ensemble trainable network with a mutual learning strategy based on a new regularization term, to model the interaction between visual and textual features learned across image and text modalities throughout the training stage. The conventional mutual learning strategy aims to encourage collaborative learning between modalities, allowing image and text modalities to simultaneously learn their discriminant features in a mutual learning manner. \textcolor{black}{The aim of introducing this approach is to enable the current modality in process to mimic the other modality by minimizing the difference in class probabilities produced by the image modality and those produced by the text modality}. However, rather than the conventional distillation-based teacher-student approach with one-way knowledge transfer from a pre-trained teacher to a student~\cite{hinton2015distilling}, the conventional mutual learning strategy starts with a pool of untrained students in a student-to-student peer-teaching model, to learn to solve the tasks collaboratively~\cite{zhang2018deep}. It turns out that conventional mutual learning achieves better results than independent learning in either a supervised or a conventional distillation learning approach from a larger pre-trained teacher. Nonetheless, conventional mutual learning is a bi-directional knowledge transfer-based method, in which the current student modality can learn from a better example from the other modality, meanwhile the good student learns from the worse modality. That is to say, if the other student is worse than the current student, the negative knowledge will be introduced and might weaken the ongoing training. This violates the motivation of the conventional mutual learning.
Thus, we introduce a mutual learning approach based on a truncated-Kullback–Leibler divergence regularization term (Tr-KLD$_{Reg}$). This approach enables the current modality to learn only the positive knowledge from the other modality and prevent the negative knowledge to be introduced in the ongoing learning of the current modality. The proposed collaborative mutual learning approach with regularization improves the quality of the final predictions of the single-modal and multi-modal modalities, and helps to overcome the drawback of the conventional mutual learning trained with the standard Kullback–Leibler divergence (KLD).

Furthermore, as one of the goals of this paper is to combine image and text features through a better multi-modal feature fusion methodology, we introduce a self-attention-based feature fusion module that serves as a middle block in our ensemble trainable network. 
Therefore, we aim to simultaneously extract more powerful and meaningful features from different middle blocks of the image and text modalities through the self-attention-based feature fusion module. This approach enables to focus more on the salient parts of feature maps of each modality, and aims to capture relevant semantic information between the pairs of image regions and text words. 
Such self-attention-based modules have recently become an elemental component in many multi-modal tasks such as visual question answering, image captioning, image-text matching, etc~\cite{NEURIPS2018_96ea64f3, Nguyen2018ImprovedFO, Yan2020ImageCV, Lee2018StackedCA}. Moreover, we adopt an early average ensemble fusion scheme in the final model to ensure a more stable and better-performing solution for the task of document image classification.

This work builds on our previous works on multi-modal networks for document image classification~\cite{souhailbakkali, 9191268}. For the rest of the paper, we denote mutual learning trained with the standard (KLD) as ML$_{KLD}$, mutual learning trained with regularization as ML$_{{Tr-KLD}_{Reg}}$, and ensemble self-attention-based mutual learning with regularization as EAML$_{{Tr-KLD}_{Reg}}$.
Following are the main contributions in this paper:

\begin{itemize}
    \item We introduce a mutual learning with a regularization term to overcome the drawback of the conventional mutual learning. This approach allows the current modality to learn the positive knowledge from the other modality instead of the negative knowledge which weakens the learning capacity of the current modality in process.
    
    \item We present a self-attention-based feature fusion module for a better multi-modal feature extraction to perform fine-grained document image classification. Our proposed self-attention-module enhances the overall accuracy of the ensemble network and achieves state-of-the-art classification performance compared to single-modal and multi-modal learning methods. 
    
    \item We perform a comprehensive ablation study on the benchmark RVL-CDIP and Tobacco-3482 datasets to analyze the effectiveness of our proposed ensemble trainable network with/without the mutual learning approach, and with/without the self-attention-based feature fusion module.
    
    \item We evaluate the performance and the generalization ability of the proposed ensemble network through inter-dataset and intra-dataset evaluation on the benchmark RVL-CDIP and Tobacco-3482 datasets for the single-modal and multi-modal fusion modalities.
\end{itemize}

The remainder of this paper is organized as follows. First, Section~\ref{sec:Related_Work}. reviews the related works and Section~\ref{sec:Architecture_Overview}. introduces our proposed architecture network. Then, Section~\ref{sec:roposed Method}. provides the details of our proposed method. We detail the experimental setup in Section~\ref{sec:Experimental Setup}. We then perform the experiments and the ablation study in Section~\ref{sec:Experiments and Ablation Study}. Finally, we give a conclusion of this paper and provide the future work in Section~\ref{sec:Conclusion and Future Work}.


\begin{figure*}[ht]
\centering
  \centerline{\includegraphics[width=\linewidth]{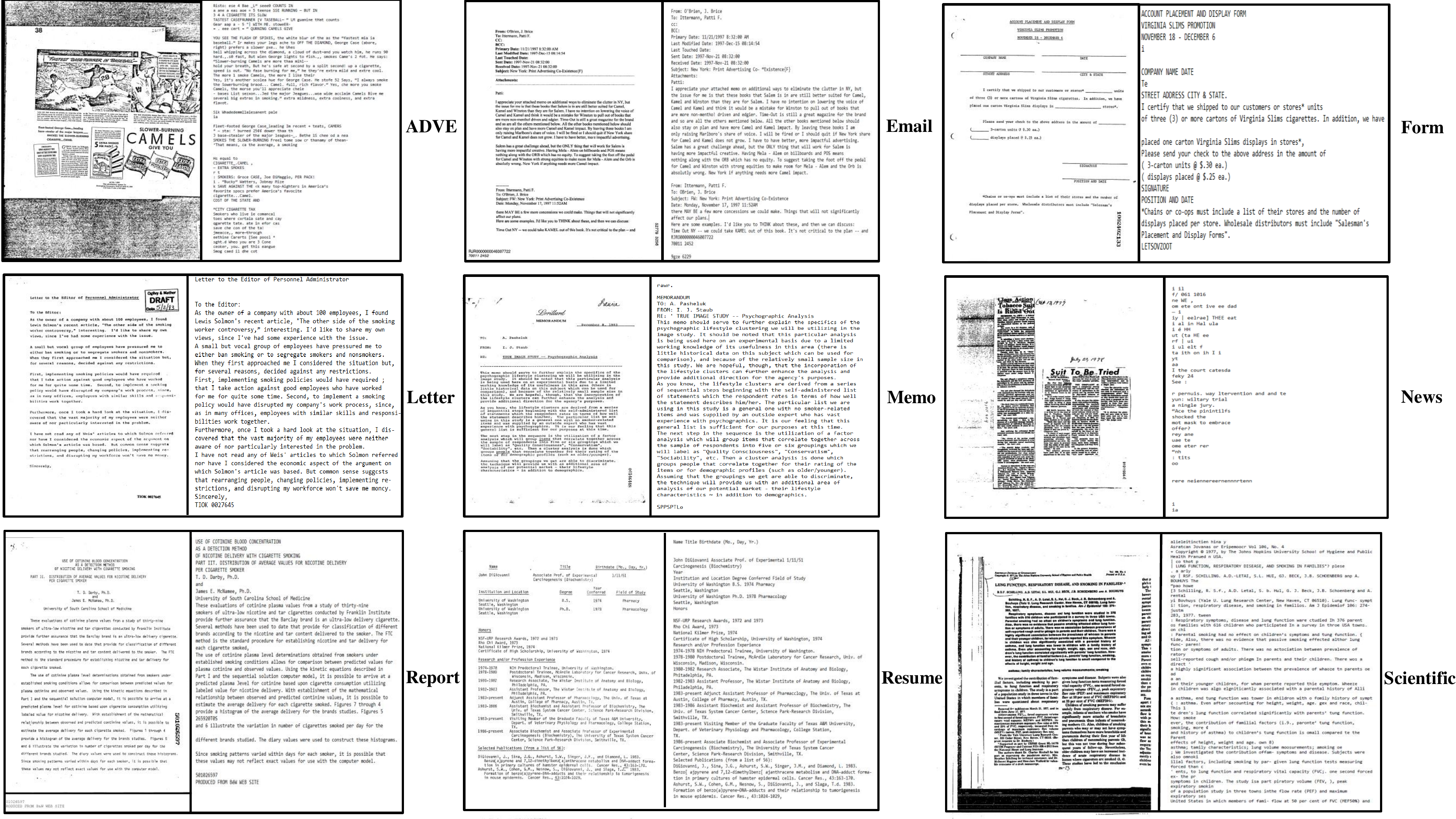}}
    \caption{\textcolor{black}{Sample images and their corresponding OCR results of 9 classes of the Tobacco-3482 dataset that overlap with the RVLCDIP dataset.}}
    \label{fig:tob_ocr}
\smallskip
\end{figure*}


\section{Related Work}
\label{sec:Related_Work}

\subsection{Image Embeddings}

Over the past few years, a variety of research studies have been proposed for document image classification. Due to the different manners of organizing each document, document images might be classified based on their heterogeneous visual structural properties and/or their textual content. Earlier attempts have utilized layout structure to convert printed documents into a complementary logical structure~\cite{Dengel1995ClusteringAC}. Region-based analysis techniques have shown notable performance in visually identifying document components, assuming that documents share a particular spatial configuration~\cite{Byun2000FormCU}. Amongst all, DCNNs-based approaches outperformed hand-crafted feature methods for the task of document image classification. Hao \etal~\cite{Hao2016ATD} proposed a novel method for table detection in PDF documents based on CNNs. Harley \etal~\cite{Harley2015EvaluationOD} proposed an alternative strategy to learn visual features through region-based approaches. Still, many pre-trained DCNNs-based approaches such as AlexNet, VGG-16, GoogLeNet, and ResNet-50~\cite{Krizhevsky2017ImageNetCW, Simonyan2015VeryDC, Szegedy2015GoingDW, He2016DeepRL, Afzal2017CuttingTE, tensmeyer2017analysis, Klsch2017RealTimeDI} have been used along transfer learning to achieve accurate document image classification results on the RVL-CDIP\footnote{\url{https://www.cs.cmu.edu/~aharley/rvl-cdip/}} and Tobacco-3482 datasets.

\begin{figure*}[ht]
\centering
  \includegraphics[width=\linewidth]{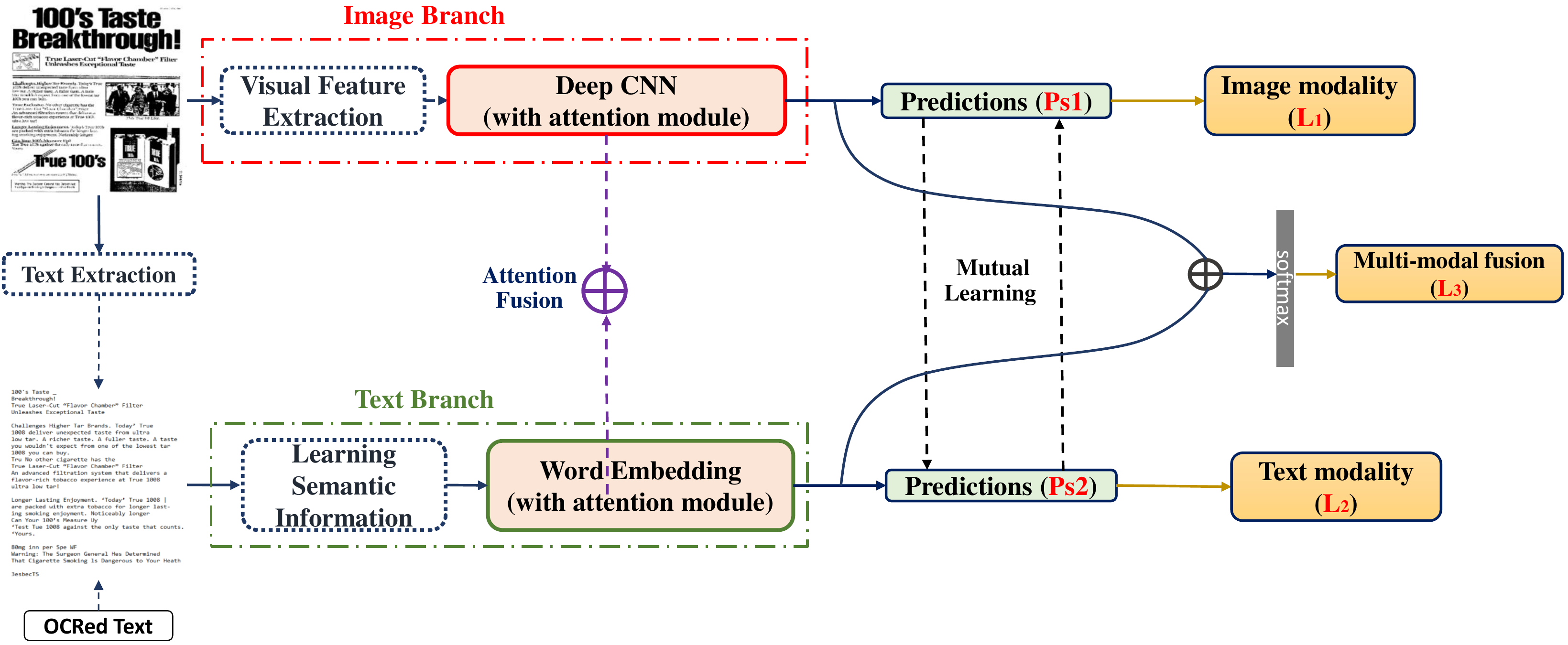}
  \caption{The proposed Ensemble Self-Attention-based Mutual Learning Network (EAML$_{{Tr-KLD}_{Reg}}$)}
    \label{fig:DAPML_architecture}
\end{figure*}

\subsection{Text Embeddings}

Recently, classifying textual content extracted from document images have been also investigated. In many natural language processing (NLP) tasks, the representation of words has drawn significant attentions. The development of static word embeddings such as Word2Vec, Glove~\cite{Mikolov2013EfficientEO, Pennington2014GloveGV}, to contextualized dynamic word embeddings such as ELMO, Fasttext, XLNet, and Bert~\cite{Peters2018DeepCW, Yang2019XLNetGA, Mikolov2018AdvancesIP, Devlin2019BERTPO} have made a huge progress to address the polysemy problem and the semantic aspect of words. In the meantime, several approaches handled the task of document image classification by performing optical character recognition (OCR) techniques. Yang \etal~\cite{Yang2017LearningTE} combined generated text features with visual features in a fully convolutional neural network. Also, \cite{Augereau2014ImprovingCO, Dauphinee2019ModularMA} experimented with shallow Bag-of-Words (BoW) along visual features in a two-modality classifier. Moreover, similar to our approach, Lai \etal~\cite{Lai2015RecurrentCN} presented a hybrid approach to extract contextual information using a RNN-CNN.


\subsection{Multi-Modal Embeddings}
As stated before, documents are natively multi-modal. Multi-modal learning for computer vision and natural language processing has been widely used for image and text level understanding problems such as text document image based classification, visual question answering~\cite{Zhou2015SimpleBF, Yang2016StackedAN}, image captioning~\cite{8578734} and image-text matching~\cite{Li2019VisualSR}. Most multi-modal fusion and attention learning methods require multi-modal reasoning over multi-modal inputs that are represented into a common space, where data related to the same topic of interest tend to appear together. For the multi-modal fusion methods, earlier attempts used naive concatenation, element-wise multiplication, and/or ensemble methods for multi-modal features~\cite{Gallo2018ImageAE, Sierra2018CombiningTA, zahavy2016picture, Yang2019ExploringDM}. Latest works like \cite{audebert2019multimodal, Asim2019TwoSD} introduced multi-modal deep networks that jointly learn visual and textual features through a fusion methodology. Noce \etal~\cite{Noce2016EmbeddedTC} proposed an approach that combines OCR and NLP algorithms to extract and manipulate relevant text concepts from document images, which are visually embedded within each document image to improve the classification results of a convolutional neural network. Fukui \etal~\cite{Fukui2016MultimodalCB} proposed a multi-modal compact bi-linear pooling to efficiently and expressively combine multi-modal features. Xu \etal~\cite{Xu2020LayoutLMPO, Xu2020LayoutLMv2MP} have recently proposed a novel architecture to merge textual and layout information for document image classification. 
\textcolor{black}{Finally, Souhail \etal~\cite{souhailbakkali, 9191268} proposed a multi-modal learning network that jointly learns the features of image and text modalities through different fusion schemes. The proposed methods have shown a superior performance compared to the single modalities, and thus, have achieved state-of-the-art performance on the RVL-CDIP and Tobacco-3482 datasets using heavyweight and lightweight deep neural networks along with different static and dynamic word embeddings.}

\subsection{Self-Attention-based Fusion Embeddings}

The attention learning was adopted to learn to attend to the most relevant regions of the input space in order to assign different weights to different regions. It was first proposed by Bahdanau \etal~\cite{Bahdanau2015NeuralMT} for neural machine translation. The mechanism is firstly used for machine translation where the most relevant words for the output often occur at similar positions in the input sequence. \textcolor{black}{Later, Vaswani \etal~\cite{vaswani2017attention} proposed a self-attention module in machine translation models which could achieve state-of-the-art results at the moment. Then, the self-attention module was introduced to guide the visual attention from images.}
For the image modality, the self-attention-based modules learn to focus on particular image regions within a given document image~\cite{Wang2018NonlocalNN, ramachandran2019standalone, Zhao2020ExploringSF}. Beyond the visual attention modules that are applied solely to the image modality, recent studies have introduced co-attention models that learn simultaneously from visual and textual attention to benefit from fine-grained representations of both modalities~\cite{NEURIPS2018_96ea64f3, Nguyen2018ImprovedFO}.
Wang \etal~\cite{wang2019position} proposed a novel position-focused attention network to investigate the relation between the visual and textual views. Chen \etal~\cite{chen2016abccnn} proposed a question-guided attention map that projects the question embeddings to the visual space, and formulates a configurable convolutional kernel to search the image attention region. Furthermore, some existing works that handled the task of jointly learning the interaction between image and text features used co-attention and self-attention modules~\cite{lu2017hierarchical, Yu_2017_ICCV, Yu2019MultimodalUA, Yu_2018}. 

\section{Architecture Overview}
\label{sec:Architecture_Overview}

\textcolor{black}{The proposed ensemble deep network (see Figure.~\ref{fig:DAPML_architecture}) is based on a multi-modal architecture, which consists of the image, text, and image/text fusion modalities. The image and text modalities are dedicated to extract visual features and textual embeddings respectively. The fusion branch is used to combine the extracted image and text features into multi-modal features. After the training of the ensemble network, the classification of document images is conducted by either the image modality or the text modality. Moreover, the visual features and the the text embeddings learned are fused to conduct document image classification in a multi-modal manner.}

\subsection{Image Modality}
The image modality extracts the visual features using the Inception-ResNet-V2 \cite{Szegedy2017Inceptionv4IA} as a backbone network, which is a convolutional neural network that achieved state-of-the-art results on the ILSVRC image classification benchmark. The model has 54.36 M parameters. 

\subsection{Text Modality}
Further, we process all document images with an off-the shelf optical character recognition (OCR) system, \ie Tesseract OCR\footnote{\url{https://github.com/tesseract-ocr/tesseract}} to extract the text from document images. \textcolor{black}{Since the document images from RVLCDIP and Tobacco-3482 datasets are well-oriented and relatively clean, it is quite straightforward to run the Tesseract OCR engine on such documents. We utilized this OCR engine to conduct a fully automatic page segmentation without orientation or script detection. We analyzed the output of OCR and find a lot of errors in the recognition especially for the classes (Handwritten, Notes), due to its incapability to recognize handwriting. Besides, the tesseract OCR engine is not always good at analyzing the natural reading order of documents. \textcolor{black}{For example, it may fail to recognize that a document contains two columns, and may try to join text across columns, which is the case of some samples from the classes (ADVE, Scientific) as shown in the qualitative results of the OCR engine in the Figure.~\ref{fig:tob_ocr}. In addition, it may produce poor quality OCR results, as a result of poor quality scans, or the distinct forms of document images as the sample shown in the Figure.~\ref{fig:tob_ocr} which corresponds to the class (News). They may contain handwritten text, tables, figures, and multi-column layouts}. The embedded features extracted from the generated text corpus are computed using Bert-base model \cite{Devlin2019BERTPO}. It is a contextualized bi-directional word embedding mechanism, that joints word representation conditioned on both left and right context in all layers using self-attention-based approaches.}

\subsection{Multi-Modal Module}
After the training of the image modality/branch and the text modality/branch by the proposed mutual learning approach with regularization (\textit{i.e.} ML$_{{Tr-KLD}_{Reg}}$), we attempt to fuse these two modalities/branches to simultaneously learn the image and text features extracted from the two image and text branches. Moreover, we adopt an early fusion methodology, (\textit{i.e.} average ensembling) as in~\cite{souhailbakkali}, which enables to enhance the global performance of multi-modal networks.

\subsection{Self-Attention-based Fusion Module}
The proposed self-attention-based fusion module has been inspired by the attention modules in the squeeze and excitation network~\cite{Hu_2018_CVPR}, which is based on re-weighting the channel-wise responses in a certain layer of a CNN by using soft self-attention in order to model the inter-dependencies between the channels of the convolutional features. As shown in Figure.~\ref{fig:Attention_module} (a), the attention fusion module is used as a middle fusion block in our ensemble trainable network. The intermediate features extracted from the middle blocks of the image branch (\textit{e.g.} the output of Residual block0) and the text branch (\textit{e.g.} the output of Transform block0) are passed to the corresponding attention block as the inputs of the attention block. The channel-wise information is then extracted from the input image or text intermediate features by performing down-sampling with the global average pooling and global max pooling layers in the attention blocks (see Figure.~\ref{fig:Attention_module} (b)). The generated channel-wise features are then inputted to the self-attention block(s) to compute the attention maps. Specially, the self-attention maps obtained from the different self attention blocks are concatenated as the final self-attention map in the visual attention block. Finally, the obtained self-attention maps from the visual attention block and text attention block are concatenated to generate the fusion attention map of the different modalities. The obtained fusion attention map is multiplied by the image and text intermediate features respectively (\textit{i.e.} the input of the visual and text attention block) as the input to the following Residual/Transform block in the image/text branch (see Figure.~\ref{fig:Attention_module} (a)).

\section{Proposed Method}
\label{sec:roposed Method}
In this section, we detail the proposed multi-modal mutual learning and self-attention-based feature fusion approaches.

\subsection{Multi-Modal Mutual Learning}
 
As seen in the Figure~\ref{fig:DAPML_architecture}, the proposed multi-modal mutual learning network consists of three different modalities: image modality (image branch), text modality (text branch) and the multi-modal modality (fusion of the two image an text modalities). 

Consider a training dataset with a set of samples and labels $(x_n, y_n) \in (\mathcal{X}, \mathcal{Y})$, over a set of $K$ classes $\mathcal{Y} \in \{1,2,..,K\}$. To learn the parametric mapping function $f_s(x_n) : \mathcal{X} \mapsto \mathcal{Y}$, \textcolor{black}{we train our ensemble network with the parameter $f_s(x_n, {\Theta})$, where ${\Theta}$ are the parameters obtained by minimizing a training objective function $\mathcal{L}_{train}$ denoted as: 
\begin{equation}
{\Theta} = \underset{\theta}{\arg\min} \mathcal{L}_{train}(y, f_s(x, {\theta}))
\label{eq:equation10}
\end{equation}
}
The total training loss of the ensemble network $\mathcal{L}_{train}$ is the sum of the weighted losses of the different modalities, \ie the image modality loss $\mathcal{L}_1$, the text modality loss $\mathcal{L}_2$ and the multi-modal fusion (image/text) loss $\mathcal{L}_3$. Specifically, $\mathcal{L}_1$ and $\mathcal{L}_2$ are obtained by the mutual learning, which can be also called as the mutual learning loss. Thus, the total loss $\mathcal{L}_{train}$, for a pair $(x_n,y_n)$, is defined as follows:


\begin{equation}
\mathcal{L}_{train}(\mathbf{X_n};\Theta) = \sum_{i=1}^{M}w_i\mathcal{L}_i(\mathbf{X}_n^{(i)};\Theta_i) = w_1\mathcal{L}_1 + w_2\mathcal{L}_2 +w_3\mathcal{L}_3  
\label{eq:equation11}
\end{equation}
where $M = 3$ is the number of modalities to be performed. ${X_i}$ and ${\Theta_i}$ are the corresponding features and the parameters learned from each modality, $\Theta=\{\Theta_i\}_{i=1}^{M}$ are the overall parameters of the networks to be optimized by $\mathcal{L}_{train}$. $w_i \in[0,1]$ s.t. $\sum w_i=1$ denote hyper-parameters which balance the independent loss terms. \textcolor{black}{Thus $\mathbf{X}_i \in \mathbb{R}^{d_i}$, where $d_i$ is the dimension of the features $X_i$, and $\mathcal{L}_i, w_i \in \mathbb{R}^1$.}

\subsubsection{Mutual Learning Loss}

The conventional mutual learning task loss consists of two losses: a supervised learning loss (\eg cross-entropy loss) and a mimicry loss (\eg Kullback-Leibler divergence (KLD). The conventional mutual learning setting aims to help the training of the current modality by transferring the knowledge between one or an ensemble of modalities in a mutual learning manner as in~\cite{zhang2018deep}. However, the knowledge learned from the other modality through the conventional (KLD) includes both the negative part and the positive part that is transferred to the current modality. Yet, instead of using the standard (KLD) in the original mutual learning, we propose a so-called truncated-KLD loss (Tr-KLD$_{Reg}$) as a new regularization term in the training loss of the current modality, which enables to filter the negative knowledge learned from the other modality, and only keep the knowledge being positive to the current modality (see Equation.~\ref{eq:equation14}). In this work, the cross-entropy loss $\mathcal{L}_{s}$ of the current modality in process can be written as:

\begin{equation}
\begin{split}
\mathcal{L}_{s}(\mathbf{X};\Theta)
= \sum^{K}_{k=1}-y_k\log(\mathcal{P}_{s}(\hat y_k|\mathbf{X},\theta_k)) \\ 
\end{split}
\label{eq:equation12}
\end{equation}
where the probability $\mathcal{P}_{s}$ is the softmax operation given by:
 
 \begin{equation}
 \begin{split}
 \mathcal{P}_{s}(\mathbf{X};\theta_k)
 &= \frac{e^{f^{\theta_k}(\mathbf{X})}}{\sum^{K}_{k'} e^{f^{\theta_{k'}}(\mathbf{X})}}
 \end{split}
 \label{eq:equation13}
 \end{equation}
where $K$ is the number of classes in the dataset, ${y_k}$ is the one-shot label of the feature $\mathbf{X}$ of the input sample, $P_{s}$ is the class probability estimated by the softmax function.
The truncated-Kullback-Leibler divergence regularization (Tr-KLD$_{Reg}$) loss of the current modality in process $\mathcal{D}_{{KL}_{Reg}}$ is given by:

\begin{equation}
\mathcal{D}_{{KL}_{Reg}}(\mathcal{P}_{{s}_{2}}\parallel{\mathcal{P}_{{s}_{1}}}) = \sum^{K}_{k=1}\mathcal{P}_{{s}_{2}}
\max\left\{0, \log\left(\frac{\mathcal{P}_{{s}_{2}}}{\mathcal{P}_{{s}_{1}}}\right)\right\}
\label{eq:equation14}
\end{equation}
where $P_{{s}_{1}}$ is the class probability estimated by the current modality, while $P_{{s}_{2}}$ refers to the class probability estimated by the other modality. \textcolor{black}{In this way, the mutual learning approach transfers the positive knowledge learned from the current modality to the other modality, by adapting the conventional mutual learning with the constraints of the mimicry loss $\mathcal{D}_{{KL}_{Reg}}$. (\ie Tr-KLD$_{Reg}$).} In the following part, $P_{{s}_{1}}$ refers to the class probabilities of the image modality, while $P_{{s}_{2}}$ refers to the class probabilities of the text modality. \\

\begin{figure*}[htb]
\centering
\begin{minipage}{.9\textwidth}
  \centering
  \subfloat[The architecture for the attention fusion between image modality and text modality.]{
 \includegraphics[width=\linewidth]{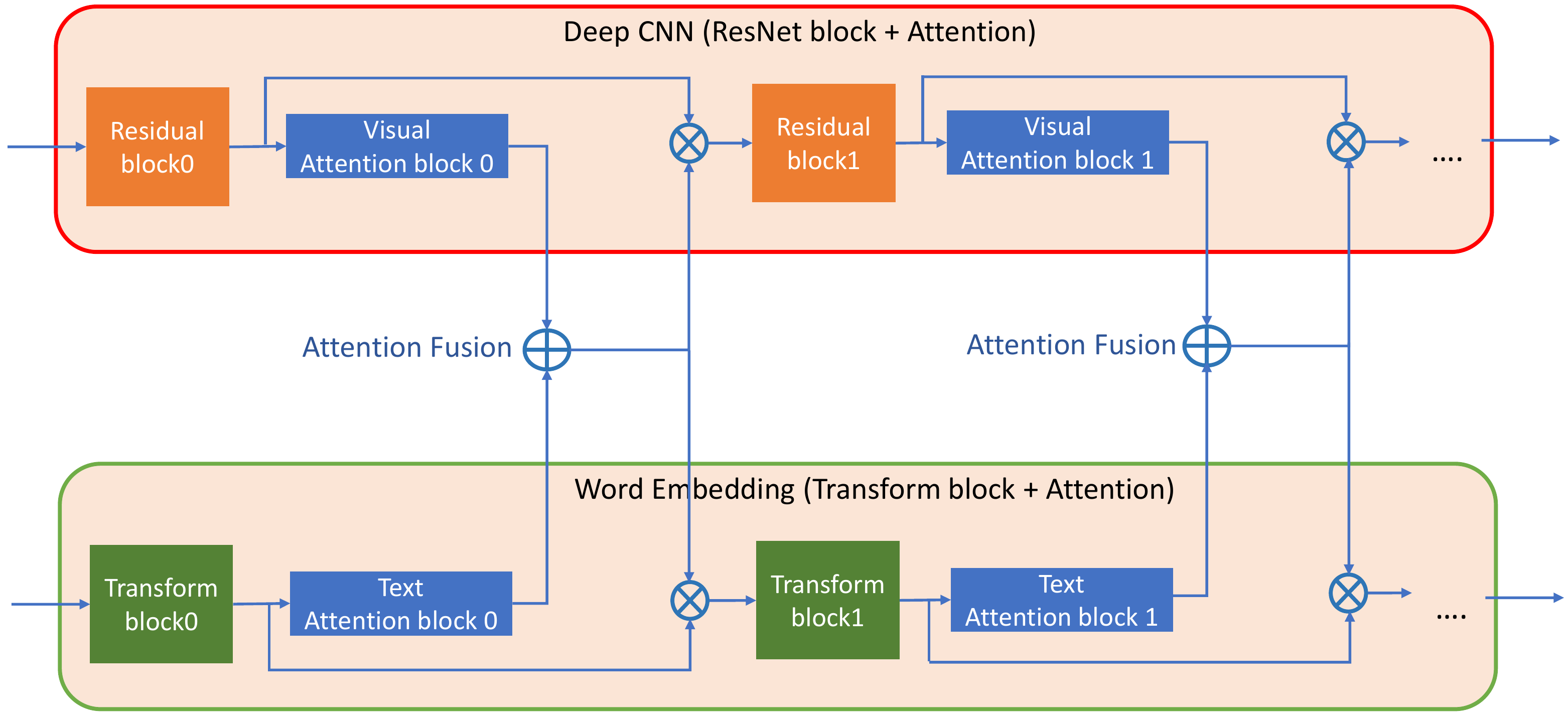}}\quad
\end{minipage}
\hspace{0.1\linewidth}
\begin{minipage}{.9\textwidth}
  \centering
  \subfloat[Visual/Textual attention block]{
  \includegraphics[width=\linewidth]{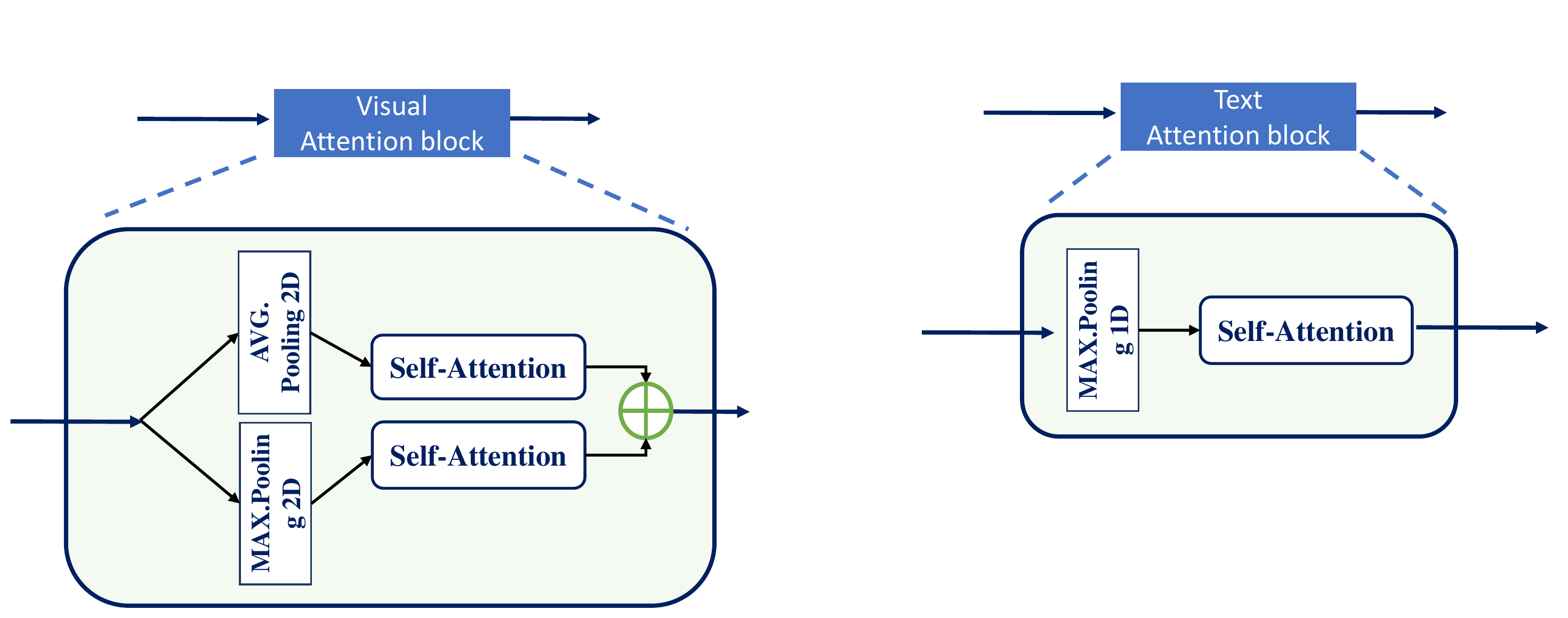}}\quad
\end{minipage}
\caption{The proposed attention-based Fusion Module}
\label{fig:Attention_module}

\end{figure*}

(i)\noindent~\textbf{Image Modality Setting:}
For the image modality, the overall loss function $\mathcal{L}_1$ is given by:    

\begin{equation}
\mathcal{L}_1(\mathbf{X}_1;\Theta_1) =
\mathcal{L}_{{s}_{1}}(\mathbf{X}_1;\Theta_1) + \\ \beta\mathcal{D}_{{KL}_{Reg}}(\mathcal{P}_{{s}_{2}}\parallel{\mathcal{P}_{{s}_{1}}})\! 
\label{eq:equation17}
\end{equation} 
where $\beta = 0.5 $ is a hyper-parameter denoting the regularization weight.

The motivation of the conventional mutual learning aims to augment the training capacity of the network, by introducing the mimicry loss to align the classification probability of the current modality to the other modality with better training. However, it is not always true that the other/text modality performs better than the current/image modality. In that case, the ongoing training of the current/image modality will be weakened by the sum of the mimicry loss with the supervised loss (\ie the cross-entropy loss for the classification of the document image). For instance, the mutual learning with regularization ${D}_{{KL}_{Reg}}$ loss will encourage the current/image modality to learn only the positive knowledge from the other/text modality, and thus, prevent the negative knowledge to be introduced in the ongoing training of the current/image modality. \\
(ii)\noindent~\textbf{Text modality Setting:}
For the text modality, the overall loss function $\mathcal{L}_2$ can be written as:  

\begin{equation}
\mathcal{L}_2(\mathbf{X}_2;\Theta_2) =
\mathcal{L}_{{s}_{2}}(\mathbf{X}_2;\Theta_2) + \\ \beta\mathcal{D}_{{KL}_{Reg}}(\mathcal{P}_{{s}_{1}}\parallel{\mathcal{P}_{{s}_{2}}})\! 
\label{eq:equation18}
\end{equation} 
Similarly to the image modality setting, the mutual learning with regularization ${D}_{{KL}_{Reg}}$ loss will prevent to transfer the negative knowledge that might be introduced from the other/image modality, and thus, will encourage to transfer only the positive knowledge to the current/text modality throughout the training process.

\subsubsection{Multi-Modal Learning Loss}
\label{sec:multi-modal feature learning}

Instead of classifying document images using the independent image or text modalities mentioned before, we can also conduct document image classification in a multi-modal manner by combining the image features and text embeddings extracted from the two modalities trained with the mutual learning approach with regularization (\ie ML$_{{Tr-KLD}_{Reg}}$).
We directly superpose the visual features of the trained image modality and text embeddings of the trained text modality to generate the ensemble cross-modal features as shown in the Equation~\ref{eq:equation20}.
Note that the dimension of the features extracted from the image modality and the text modality are equal in this work and are denoted as $d$. A softmax layer at the end of the network is used to learn the classification of document images based on the ensemble cross-modal features \textcolor{black}{$\mathbf{X}_3$. The parameter $\Theta_3$ of the softmax layer is optimized by the cross-entropy loss function $\mathcal{L}_3(\mathbf{X}_3;\Theta_3)$ which is given by:
 \begin{equation}
 \mathcal{L}_3(\mathbf{X}_3;\Theta_3)=-\sum^{K}_{k=1}y_klogP(\hat y_k|\mathbf{X}_3,\Theta_3)
 \label{eq:equation19}
 \end{equation}
 }
 \textcolor{black}{With $\mathbf{X}_3$ is given by:}
\textcolor{black}{
 \begin{equation}
 \mathbf{X}_3 = [X_1+X_2] , \quad \mathbf{X}_3 \in \mathbb{R}^{d}
 \label{eq:equation20}
 \end{equation}
}

\subsection{Self-Attention-based Fusion Module}

The aim of the self-attention-based fusion module (see Figure.~\ref{fig:Attention_module}) is to enhance the representation of the concatenated image and text feature maps to capture their salient features while eliminating to some extent the irrelevant or noisy ones. The adopted self-attention-based fusion module has been inspired by the attention module in~\cite{vaswani2017attention, Hu_2018_CVPR}, which is based on the channel-wise re-calibration of feature maps to model the dependency of channels. The intermediate feature maps of each single modality can be interpreted as a set of local descriptors that include global information in the decision process of the network. This is achieved by using global max pooling and global average pooling layers to generate channel-wise information. The advantage of these pooling operations is to enforce correspondences between feature maps and categories. 

Consider a set of input features $\mathbf{X} = [\mathrm{x}_{1},...,\mathrm{x}_{m}] \in \mathbb{R}^{{m}.{d}_{x}}$ and output features $\mathcal{F} = [\mathrm{f}_{1},...,\mathrm{f}_{m}] \in \mathbb{R}^{{m}.{d}_{f}}$, where $\mathrm{m}$ is the number of samples,  ${d}_{x}$ and ${d}_{f}$ are the dimensions of input and output features respectively. For the image modality, the input features $\mathbf{X}$ are passed to global average pooling and global max pooling layers. \textcolor{black}{The spatial information for each layer is computed as:}
\textcolor{black}{
\begin{equation}
    \mathcal{F}^{'}_{{I}_{Avg}} = GlobalAvgPool2D (\mathbf{X}_{{I}_{Avg}})
 \label{eq:equation1}
\end{equation}
}
\textcolor{black}{
\begin{equation}
    \mathcal{F}^{'}_{{I}_{Max}} = GlobalMaxPool2D (\mathbf{X}_{{I}_{Max}})  
\label{eq:equation1_1}
\end{equation}
}
where $\mathcal{F}^{'}_{{I}_{Avg}}$, and ${F}^{'}_{{I}_{Max}}$ correspond to the intermediate feature maps of the intermediate input features ${X}_{{I}_{Avg}}$, and ${X}_{{I}_{Max}}$ of the image modality.
For the text modality, the input features are fed to a global max pooling layer:

\textcolor{black}{
\begin{equation}
    \mathcal{F}^{'}_{{T}_{Max}} = GlobalMaxPool1D (\mathbf{X}_{{T}_{Max}})
    \label{eq:equation2}
\end{equation}
}
\textcolor{black}{
where ${F}^{'}_{{T}_{Max}}$ corresponds to the intermediate feature maps of the input features ${X}_{{T}_{Max}}$ of the text modality.
For our proposed self-attention-based fusion module, the intermediate feature maps of the image and text modalities extracted by the pooling operations are fed to three independent fully-connected layers which correspond to the vectors query, keys, and values respectively as follows:
\begin{equation}
\mathrm{Q} = 
\mathrm{FC}_{q}(\mathcal{F}^{'}); \\
\mathrm{K} = \mathrm{FC}_{k}(\mathcal{F}^{'}); \\ 
\mathrm{V}_{I} = 
\mathrm{FC}_{v}(\mathcal{F}^{'})
\label{eq:equation3}
\end{equation}
where $\mathrm{Q}, \mathrm{K}, \mathrm{V} \in \mathbb{R}^{{m}.{d}}$ are three vectors of the same shape used to calculate the attention function, which consists of computing the compatibility of the query with the key vectors to retrieve the corresponding value. \\
Given a query $\mathrm{q} \in \mathrm{Q}$ and all keys $\mathrm{K}$, we calculate the dot products of $\mathrm{q}$ with all keys $\mathrm{K}$, divide each by a scaling factor $\sqrt{{d}_{f}}$ and apply the softmax function to get the attention weights on the values. The output features of each self-attention module of image and text modalities $\mathcal{F}$ are given as follows:\\
\begin{equation}
\mathrm{A} = Softmax(\frac{\mathrm{Q}.{K}^{\mathcal{T}}}{\sqrt{{d}_{f}}})
\label{eq:equation5}
\end{equation}
\begin{equation}
\mathcal{F} = \mathrm{A}.\mathrm{V}
\label{eq:equation6}
\end{equation}
where $\mathrm{A}$ is the attention map containing the attention weights for all query-key pairs, and the output features of the self-attention blocks $\mathcal{F}$ are the weighted summation of the values $\mathrm{V}$ determined by the attention function $\mathrm{A}$.}

\textcolor{black}{Learning an accurate attention map $\mathrm{A}$ is crucial for self-attention learning. The scaled dot-product attention in the Equations~(\ref{eq:equation5}, \ref{eq:equation6}) models the relationship between feature pairs. Once the spatial information are extracted and fed into the self-attention blocks to compute the attention maps, they are then concatenated and multiplied by the input features of the image and text modalities for adaptive feature fusion, which is computed as follows:
\begin{equation}
    \mathcal{M}(\mathcal{F}) = \sigma(\mathcal{F}). \mathcal{F} 
    \label{eq:equation9}
\end{equation}
Where $\mathcal{M}$ is the feature map that is passed to the following intermediate image and text blocks of the image and text modalities. The term $\sigma(.)$ denotes the sigmoid function. This feature map generated by the proposed self-attention-based fusion module focuses on the important features of the channels and concentrates on where the salient features are located.
}
\setlength\tabcolsep{3 pt}
\begin{table*}[tbh]
\small
\centering
\caption{The overall classification accuracy(Acc.), recall(R.), precision(Pr.) metrics of the proposed approaches on the RVL-CDIP dataset.}
\resizebox{\textwidth}{!} {%
    \begin{tabular}{lcccccccccccc}
    \hline\noalign{\smallskip}
     \multicolumn{1}{c}{} && \multicolumn{9}{c}{Modality} \\
         \noalign{\smallskip}\hline\noalign{\smallskip}
         \multicolumn{1}{c}{Method} && \multicolumn{3}{c}{Image Modality} && \multicolumn{3}{c}{Text Modality} && \multicolumn{3}{c}{Multi-modal Fusion} \\
         \noalign{\smallskip}\hline\noalign{\smallskip}
          && Acc.(\%) & R. & Pr. && Acc.(\%) & R. & Pr. && Acc.(\%) & R. & Pr.\\
         \cmidrule{3-5}\cmidrule{7-9}\cmidrule{11-13}
         Independent Learning (IL) && 85.04 & 0.85 & 0.85 && 84.96 & 0.85 & 0.85 && 94.44 & 0.94 & 0.94 \\
         Mutual Learning (ML$_{KLD})$ && 88.87 & 0.89 & 0.88 && 80.89 & 0.81 & 0.80 && 90.06 & 0.90 & 0.90 \\
         Mutual Learning (ML$_{{Tr-KLD}_{Reg}}$) && 90.81 & 0.91 & 0.91 && 88.80 & 0.89 & 0.89 && 96.28 & 0.96 & 0.96\\
         Ensemble Self-Attention Mutual Learning (EAML$_{{Tr-KLD}_{Reg}}$) && \textbf{97.67} & 0\textbf{.98} & \textbf{0.98} && \textbf{97.63} & \textbf{0.98} & \textbf{0.98} && \textbf{97.70} & 0\textbf{.98} & \textbf{0.98} \\
    \noalign{\smallskip}\hline
    \end{tabular}%
    }
    \label{tab:proposedmethods}
\end{table*}

\section{Experimental Setup}
\label{sec:Experimental Setup}
\subsection{Datasets}

To evaluate the performance of our proposed ensemble trainable network presented in the Section~\ref{sec:Architecture_Overview}, two benchmark datasets have been used. First, we introduce a subset of the IIT-CDIP Test Collection known as RVL-CDIP. This dataset consists of gray-scale labeled scanned document images into $16$ classes (advertisement, budget, email, file folder, form, handwritten, invoice, letter, memo, news article, presentation, questionnaire, resume, scientific publication, scientific report, specification). The dataset is split into training set which contains $320,000$ images, and  a validation and a test sets which contain $40,000$ images each. Some representative images from the dataset are shown in the Figure~\ref{fig:dataset}.
Secondly, we use the public Tobacco-3482 dataset to evaluate the performance and the generalization ability of the ensemble trained network on the common classes between the two datasets. The Tobacco-3482 dataset contains $3,482$ gray-scale document images of 10 categories: ADVE, Email, Form, Letter, Memo, News, Notes, Report, Resume and Scientific. 

\subsection{Preprocessing}

As the image modality require as an input, images of a fixed size, we first downscale all images to 229 x 229 pixels. Intuitively, when training DCNNs, data augmentation has shown to be effective for real-world image classification~\cite{Krizhevsky2017ImageNetCW}. The training data is augmented by shifting it horizontally and vertically with a range of 0.1. Also, shear transform is applied with a range of 0.1. To improve regularization of our image modality, cutout~\cite{devries2017improved} is applied, which augments the training data by partially occluded versions of the existing sample images.
On the other hand, document images from the RVL-CDIP dataset are well-oriented and relatively clean. Hence, we run the Tesseract OCR engine. We used the version 4.0.0-beta.1 of Tesseract based on a LSTM engine to aim for better accuracy. The resulting extracted text was not post-processed. Although document information might be lost in OCR, such as typeface, graphics, layout, stop words, mis-spellings, symbols and characters. it could benefit from some level of spell checking to improve the semantic learning. However, we chose to provide the true output of Tesseract OCR as it is.

\subsection{Training details} 

The network used in our proposed approaches were conducted on a 4 NVIDIA RTX-2080 GPU, using stochastic gradient descent optimizer (SGD), with Nesterov momentum, mini-batch size of 16, and a learning rate of 1e-3 decayed with a value of 0.5 every 10 epochs. the learning rate decay is defined as :
\begin{equation}
    lr = initial\_lr * drop^{\left(\frac{iter}{iter\_drop} \right)}
    \label{eq:equation21}
\end{equation}
The mutual learning strategy with regularization (\ie ML$_{{Tr-KLD}_{Reg}}$) is performed in each mini-batch throughout the training process. At each iteration, the predictions of each modality are computed and the parameters are updated according to the predictions of the other modality as in the Equations.~(\ref{eq:equation17}, \ref{eq:equation18}, \ref{eq:equation19}). The optimization process of parameters $\Theta_1$, $\Theta_2$, and $\Theta_3$ is performed iteratively until convergence. We considered early stopping within 10 epochs to stop the training process once the model's performance stops improving on the hold out validation dataset.
\begin{table}[tbh]
\small
\centering
\caption{The overall classification accuracy of our best EAML$_{{Tr-KLD}_{Reg}}$ method against baseline methods on the RVL-CDIP dataset.}
    \begin{tabular}{lcc}
    \hline\noalign{\smallskip}
     Method & Model & Accuracy(\%)\\
     \noalign{\smallskip}\hline\noalign{\smallskip}
    Image & \multirow{3}{*}{Nicolas \etal \cite{audebert2019multimodal}} & 89.1\\
    Text && 74.6\\
    Multi-modal && 90.6\\
    \noalign{\smallskip}\hline\noalign{\smallskip}
    Image & \multirow{3}{*}{Dauphinee \etal \cite{Dauphinee2019ModularMA}} & 90.24\\
    Text && 82.23\\
    Multi-Modal && 93.07\\
    \noalign{\smallskip}\hline\noalign{\smallskip}
    Image & \multirow{3}{*}{Cross-Modal \cite{souhailbakkali}} & 91.45\\
    Text && 84.96\\
    Multi-modal && 97.05\\
    \noalign{\smallskip}\hline\noalign{\smallskip}
    Image & \multirow{3}{*}{EAML$_{{Tr-KLD}_{Reg}}$ (Ours)} & \textbf{97.67}\\
    Text && \textbf{97.63}\\
    Multi-Modal && \textbf{97.70}\\
    \noalign{\smallskip}\hline\noalign{\smallskip}
    \multirow{8}{*}{Baselines} 
    & Harley \etal \cite{Harley2015EvaluationOD} & 89.80\\
    & Csurka \etal \cite{Csurka2016WhatIT} & 90.70\\
    & Tensmeyer \etal \cite{tensmeyer2017analysis} & 90.94\\
    & Azfal \etal \cite{Afzal2017CuttingTE} & 90.97\\
    & Das \etal \cite{8545630} & 91.11\\
    & Das \etal \cite{8545630} & 92.21\\
    & Ferrando \etal \cite{Ferrando2020ImprovingAA} & 92.31\\
    & Xu \etal \cite{Xu2020LayoutLMPO} & 94.42\\
    & Xu \etal \cite{Xu2020LayoutLMv2MP} & 95.64\\
    \noalign{\smallskip}\hline
    \end{tabular}
    \label{tab:Comparisontable}
\end{table}
\section{Experiments and Ablation Study}
\label{sec:Experiments and Ablation Study}

\subsection{Evaluation Protocol}

To evaluate the performance and the generalization ability of our proposed ensemble network, we proceed with intra-dataset and inter-dataset evaluation on the benchmark RVL-CDIP and Tobacco-3482 datasets. For the intra-dataset evaluation, we train and test the model on the same dataset to evaluate the performance of the proposed approaches. Whereas, for the inter-dataset evaluation, we train and test the ensemble network on different datasets to evaluate the generalization ability of the trained model. We first train our ensemble network on the RVL-CDIP dataset, then we employ the intra-dataset evaluation on RVL-CDIP and the inter-dataset evaluation on Tobacco-3482. Secondly, we train our ensemble network on the Tobacco-3482 dataset, then we employ the intra-dataset evaluation on Tobacco-3482 and the inter-dataset evaluation on RVL-CDIP. Note that there is no overlap between training set and test set either in intra-dataset or inter-dataset evaluation. 

We report the accuracy, recall, and precision metrics achieved on the test set for the following methods: the Independent Learning based on the single-modal image and text modalities, the Mutual Learning trained with the standard Kullback-Leibler divergence (KLD). The Mutual Learning trained with the truncated-Kullback-Leibler divergence regularization (Tr-KLD$_{Reg}$) loss, and the Ensemble Self-Attention Mutual Learning trained with (Tr-KLD$_{Reg}$), that are denoted respectively as IL, ML$_{KLD}$, ML$_{{Tr-KLD}_{Reg}}$, and EAML$_{{Tr-KLD}_{Reg}}$ (see Table.~\ref{tab:proposedmethods}). 
We also compute the average precision (AP) from prediction scores which summarizes a precision-recall curve as the weighted mean of precision achieved at each threshold, with the increase in recall from the previous threshold used as the weight:
\begin{equation}
    \mathrm{AP} = \sum_{n}(\mathrm{R}_{n} - \mathrm{R}_{n-1})\mathrm{P}_{n}
    \label{eq:equation22}
\end{equation}
where $\mathrm{P}_{n}$ and $\mathrm{R}_{n}$ are the precision and recall at the $n^{th}$ threshold. The high area under the (AP) curve represents both high recall and high precision, where high precision relates to a low false positive rate, and high recall relates to a low false negative rate. High scores for both precision and recall show that the model is returning accurate results (high precision), as well as returning a majority of all positive results (high recall). 
In addition, we compare our work against other state-of-the-art methods on the RVL-CDIP and Tobacco-3482 datasets. \textcolor{black}{Note that the baseline methods in the Tables~(\ref{tab:Comparisontable}, \ref{tab:ComparisontableonTobacco}) are not necessarily based on image and text modalities. For example, \cite{Xu2020LayoutLMPO} leverages image features to incorporate words' visual information into LayoutLM for document-level pre-training. Also, \cite{Xu2020LayoutLMv2MP} leverages pre-training text, layout and image in a multi-modal framework by using text-image alignment and text-image matching tasks in the pre-training stage, where the cross-modality interaction is better learned.}

\setlength\tabcolsep{3 pt}
\begin{table*}[tbh]
\small
\centering
\caption{The overall classification accuracy(Acc.), recall(R.), precision(Pr.) metrics of the proposed approaches on the Tobacco-3482 dataset.}
\resizebox{\textwidth}{!} {%
    \begin{tabular}{lcccccccccccc}
    \hline\noalign{\smallskip}
     \multicolumn{1}{c}{} && \multicolumn{9}{c}{Modality} \\
         \noalign{\smallskip}\hline\noalign{\smallskip}
         \multicolumn{1}{c}{Method} && \multicolumn{3}{c}{Image Modality} && \multicolumn{3}{c}{Text Modality} && \multicolumn{3}{c}{Multi-modal Fusion} \\
         \noalign{\smallskip}\hline\noalign{\smallskip}
          && Acc.(\%) & R. & Pr. && Acc.(\%) & R. & Pr. && Acc.(\%) & R. & Pr.\\
         \cmidrule{3-5}\cmidrule{7-9}\cmidrule{11-13}
         Independent Learning (IL) && 96.17 & 0.96 & 0.96 && 96.02 & 0.96 & 0.95 && 96.95 & 0.97 & 0.97 \\
         Mutual Learning (ML$_{KLD}$) && 93.69 & 0.92 & 0.92 && 88.82 & 0.87 & 0.86 && 94.84 & 0.95 & 0.93 \\
         Mutual Learning (ML$_{{Tr-KLD}_{Reg}}$) && 97.70 & 0.97 & 0.96 && 96.27 & 0.95 & 0.96 && 98.28 & 0.97 & 0.98\\
         Ensemble Self-Attention Mutual Learning (EAML$_{{Tr-KLD}_{Reg}}$) && \textbf{97.99} & \textbf{0.97} & \textbf{0.98} && \textbf{96.27} & \textbf{0.95} & \textbf{0.96} && \textbf{98.57} & \textbf{0.98} & \textbf{0.98} \\
    \noalign{\smallskip}\hline
    \end{tabular}%
    }
    \label{tab:ResultsonTobacco}
\end{table*}

\subsection{Intra-dataset Evaluation}

\subsubsection{Results on the RVL-CDIP Dataset}

On the large-scale RVL-CDIP dataset, all of the adopted approaches in this work achieve comparable performance with the state-of-the-art models. We report the overall accuracy results in the Table~\ref{tab:Comparisontable}. compared to our latest work\cite{souhailbakkali} and other baseline methods. The proposed EAML$_{{Tr-KLD}_{Reg}}$ model achieves the best performance in terms of accuracy for the single-modal image and text modalities, and for the multi-modal fusion modality at an accuracy of $97.67\%$, $97.63\%$, and $97.70\%$ respectively. \textcolor{black}{The adopted self-attention-based fusion module has shown its effectiveness in capturing simultaneously the inter-modal interactions between image features and text embeddings, along with the mutual learning approach with regularization (\ie ML$_{{Tr-KLD}_{Reg}}$)}.
Therefore, it improves the global classification performance of the single-modal and multi-modal modalities and outperforms the state-of-the-art methods.


\subsubsection{Evaluation of the Single-Modal Tasks on the RVL-CDIP dataset}

(i.)\noindent~\textbf{IL vs ML$_{KLD}$:}
The reported results in the Table~\ref{tab:proposedmethods} illustrate the impact of training the independent image and text modalities in a mutual learning manner, on the learning process of both modalities. We observe that the ML$_{KLD}$ method improves the classification performance of the image modality from $85.04\%$ to $88.87\%$, while it deteriorated the performance of the text modality from $84.96\%$ to $80.89\%$. We explain this performance deterioration of the text modality by learning the negative knowledge from the image modality. In fact, the knowledge transferred via the standard (KLD) loss harms the ongoing training of the current/text modality in process. Here, given image features from an image sample with its corresponding text embeddings, the negative learning comes from the low class probabilities predicted by the image modality, while at the same time, the text modality has made the right predictions from the same sample.
In this way, the mutual training is harmed for the text modality and its loss variation $\mathcal{L}_{2}(\mathbf{X}_2;\Theta_2)$ becomes slower. Thus, using the Mutual Learning ML$_{KLD}$ method actually makes the text modality worse than the Independent Learning (IL) method. 

Nonetheless, for the image modality, the classification accuracy has improved. \textcolor{black}{This means that transferring the knowledge from the text modality to the image modality by learning mutually from the text predictions is effective}.
\\
(ii.)\noindent~\textbf{IL vs ML$_{{Tr-KLD}_{Reg}}$:}
The classification results in the Table.~\ref{tab:proposedmethods} show that, training the image and text modalities in a mutual learning manner --trained with the regularization term (\ie Tr-KLD$_{Reg}$)-- provide an improvement compared to the IL and the ML$_{KLD}$ methods. It improves the classification accuracy of the image modality from $85.04\%$ for the IL method to $90.81\%$ for the ML$_{{Tr-KLD}_{Reg}}$ method. Also, it enhances the predictions of the text modality from $84.96\%$ to $88.80\%$ respectively.

Accordingly, the network keeps learning only from its cross-entropy loss $\mathcal{L}_{s}(\mathbf{X};\Theta)$ when the knowledge to be transferred from the other modality will harm the ongoing training of the current modality.
\\
(iii.)\noindent~\textbf{ML$_{{Tr-KLD}_{Reg}}$ vs EAML$_{{Tr-KLD}_{Reg}}$:}
The proposed self-attention-based fusion module for image and text feature fusion focuses on the salient feature maps generated from the image and the text modalities and suppresses the unnecessary ones to efficiently leverage these two modalities. \textcolor{black}{The introduction of this attention module to fuse the two modalities along with the mutual learning approach has shown its efficiency compared to the ML$_{{Tr-KLD}_{Reg}}$ method as shown in the Table.~\ref{tab:proposedmethods}}. We demonstrate that the EAML$_{{Tr-KLD}_{Reg}}$ method outperforms ML$_{{Tr-KLD}_{Reg}}$ method with a significant margin at an accuracy of $97.67\%$, $97.63\%$ for the image and text modalities respectively. The attention module enhances the classification performance of all classes for the single-modal modalities. therefore, leveraging both modalities to one another in a middle fusion manner along with the mutual learning strategy encourage collaborative learning during the training stage. 


\subsubsection{Evaluation of the Multi-Modal Tasks on the RVL-CDIP dataset}

In the multi-modal learning task, the learned image and text features are combined to conduct document image classification. At first, from the Table.~\ref{tab:proposedmethods}, we see that the multi-modal fusion predictions outperform the independent predictions of the single-modal modalities for each method. 
Moreover, jointly learning both modalities in an ensemble network benefit from training image modality and text modalities both independently (IL) and in a mutual learning manner (ML$_{{Tr-KLD}_{Reg}}$). The ensemble predictions learned across the EAML$_{{Tr-KLD}_{Reg}}$ method with an accuracy of $97.70\%$, outperform the predictions learned from training the ensemble network across either the ML$_{{Tr-KLD}_{Reg}}$, the ML$_{KLD}$, or the IL approaches at an accuracy of $96.28\%$, $90.06\%$, and $94.44\%$ respectively. That is to say, the ability of the self-attention-based fusion module along with the mutual learning strategy --trained with the regularization term (\ie Tr-KLD$_{Reg}$)-- to improve ensemble models is beneficial for the task of document image classification, which outperforms the state-of-the-art results for the multi-modal task as seen in the Table~\ref{tab:Comparisontable}. 
\setlength\tabcolsep{7 pt}
\begin{table}[tbh]
\small
\centering
\caption{The overall classification accuracy of the proposed approaches against baseline methods on the Tobacco-3482 dataset.}

    \begin{tabular}{lcc}
    \hline\noalign{\smallskip}
     Method & Model & Accuracy(\%)\\
     \noalign{\smallskip}\hline\noalign{\smallskip}
    Image & \multirow{3}{*}{Nicolas \etal \cite{audebert2019multimodal}} & 84.5\\
    Text && 73.8\\
    Multi-modal && 87.8\\
    \noalign{\smallskip}\hline\noalign{\smallskip}
    Image & \multirow{3}{*}{Asim \etal \cite{Asim2019TwoSD}} & 93.2\\
    Text && 87.1\\
    Multi-modal && 95.8\\
    \noalign{\smallskip}\hline\noalign{\smallskip}
    Image & \multirow{3}{*}{Ferrando \etal \cite{Ferrando2020ImprovingAA}} & 94.04\\
    Text && -\\
    Multi-modal && 94.90\\
    \noalign{\smallskip}\hline\noalign{\smallskip}
    Image & \multirow{3}{*}{Cross-Modal \cite{9191268}} & 96.25\\
    Text && \textbf{97.18}\\
    Multi-modal && \textbf{99.71}\\
    \noalign{\smallskip}\hline\noalign{\smallskip}
    Image & \multirow{3}{*}{EAML$_{{Tr-KLD}_{Reg}}$ (Ours)} & \textbf{97.99}\\
    Text && 96.27\\
    Multi-modal && 98.57\\
    \noalign{\smallskip}\hline\noalign{\smallskip}
    \multirow{5}{*}{Baselines} 
    & Kumar \etal \cite{Kumar2014StructuralSF} & 43.8\\
    & Kang \etal \cite{6977258} & 65.37\\
    & Afzal \etal \cite{Afzal2015DeepdocclassifierDC} & 76.6\\
    & Harley \etal \cite{Harley2015EvaluationOD} & 79.9\\
    & Noce \etal \cite{Noce2016EmbeddedTC} & 79.8\\
    \noalign{\smallskip}\hline
    \end{tabular}
    \label{tab:ComparisontableonTobacco}
\end{table}
Accordingly, The proposed EAML$_{{Tr-KLD}_{Reg}}$ method manages to correct the classification errors produced by image and text modalities during the learning process. Hence, it provides state-of-the-art classification results for the task of document image classification. 

In this manner, we showed the effectiveness of leveraging visual and textual features learned in a mutual learning with regularization strategy through a self-attention-based feature fusion module. Our approach learns simultaneously relevant and accurate information from the image modality, and the text modality during the training stage. \textcolor{black}{It enhances the ensemble model predictions by encouraging attention collaborative learning from one modality to another}. Also, it boosts the overall classification performance.
We report in the (Online Resource 1, Figures.(1,2)), the confusion matrices of the multi-modal modalities of the EAML$_{{Tr-KLD}_{Reg}}$ and the ML$_{{Tr-KLD}_{Reg}}$ methods respectively.

\subsubsection{Results on the Tobacco-3482 Dataset}

As reported in the Table.~\ref{tab:ResultsonTobacco}, which corresponds to the achieved performance on the Tobacco-3482 dataset, the EAML$_{{Tr-KLD}_{Reg}}$ method improves the classification performance significantly. 
The proposed EAML$_{{Tr-KLD}_{Reg}}$ method improves the overall performance of the single-modal and multi-modal modalities at an accuracy of $97.99\%$, $96.27\%$, and $98.57\%$ for the image modality, for the text modality, and for the multi-modal fusion modality respectively compared to other methods. Thus, it achieves compelling performance results compared to the baseline methods on the Tobacco-3482 dataset (see Table.~\ref{tab:ComparisontableonTobacco}).

\textcolor{black}{Besides, the results illustrate that training the image and text modalities in a mutual learning manner with the ML$_{KLD}$ method weakens the learning capacity of the text modality}. Therefore, we show the effectiveness of the ML$_{{Tr-KLD}_{Reg}}$ approach that transfers only the positive knowledge from the current modality in process to the other modality.

\setlength\tabcolsep{3.8 pt}
\begin{table*}[tbh]
\small
\centering
\caption{The Inter-Dataset Evaluation results of the Mutual Learning ML$_{{Tr-KLD}_{Reg}}$ method on the Tobacco-3482 dataset.}
\resizebox{\textwidth}{!} {%
    \begin{tabular}{lcccccccccccccc}
    \hline\noalign{\smallskip}
     \multicolumn{1}{c}{} && \multicolumn{10}{c}{Mutual Learning (ML$_{{Tr-KLD}_{Reg}}$)} \\
         \noalign{\smallskip}\hline\noalign{\smallskip}
         \multicolumn{1}{c}{Class Labels} && \multicolumn{3}{c}{Image Modality} && \multicolumn{3}{c}{Text Modality} && \multicolumn{3}{c}{Multi-modal Fusion} && \multicolumn{1}{c}{\#Nb. Samples}\\
         \noalign{\smallskip}\hline\noalign{\smallskip}
          && Precision & Recall & F1-Score &&  Precision & Recall & F1-Score && Precision & Recall & F1-Score &&  \\
         \cmidrule{3-13}
         Advertisement 
         && 0.9659 & 0.9659 & 0.9103
         && 0.9596 & 0.8261 & 0.8879 
         && 0.9772 & 0.9304 & 0.9532 
         && 230 \\
         Email 
         && 0.9688 & 0.9850 & 0.9768 
         && 0.9577 & 0.9833 & 0.9703
         && 0.9673 & 0.9866 & 0.9769 
         && 599 \\
         Form 
         && 0.9484 & 0.8956 & 0.9212 
         && 0.9360 & 0.8817 & 0.9080
         && 0.9408 & 0.9582 & 0.9494 
         && 431 \\
         Letter 
         && 0.8959 & 0.9718 & 0.9323
         && 0.9035 & 0.9577 & 0.9298 
         && 0.9329 & 0.9806 & 0.9561 
         && 567 \\
         Memo 
         && 0.9562 & 0.9855 & 0.9706
         && 0.9466 & 0.9726 & 0.9594
         && 0.9717 & 0.9968 & 0.9841 
         && 620 \\
         News article 
         && 0.8650 & 0.9202 & 0.8918
         && 0.8406 & 0.9255 & 0.8810
         && 0.9146 & 0.9681 & 0.9406 
         && 188 \\
         Resume 
         && 0.9836 & 1 & 0.9917
         && 0.9836 & 1 & 0.9917
         && 0.9756 & 1 & 0.9877 
         && 120 \\
         Scientific publication 
         && 0.9462 & 0.3372 & 0.4972
         && 0.8889 & 0.3372 & 0.4889
         && 0.9368 & 0.3410 & 0.50 
         && 261 \\
         Scientific report 
         && 0.2907 & 0.2491 & 0.2683 
         && 0.2707 & 0.2340 & 0.2510 
         && 0.2773 & 0.2302 & 0.2515 
         && 265 \\
    \noalign{\smallskip}\hline\noalign{\smallskip}
        Overall Accuracy (\%) && & & 84.82 &&  &  & 83.72 &&  &  & 86.68 \\
    \noalign{\smallskip}\hline
    
    \end{tabular}%
    }
    \label{tab:Evaluation of PML on tobacco}
\end{table*}


\setlength\tabcolsep{3.8 pt}
\begin{table*}[tbh]
\small
\centering
\caption{The Inter-Dataset Evaluation results of the Ensemble Self-Attention Mutual Learning (EAML$_{{Tr-KLD}_{Reg}}$) approach on the Tobacco-3482 dataset.}
\resizebox{\textwidth}{!} {%
    \begin{tabular}{lcccccccccccccc}
    \hline\noalign{\smallskip}
     \multicolumn{1}{c}{} && \multicolumn{10}{c}{Ensemble Self-Attention Mutual Learning (EAML$_{{Tr-KLD}_{Reg}}$)} \\
         \noalign{\smallskip}\hline\noalign{\smallskip}
         \multicolumn{1}{c}{Class Labels} && \multicolumn{3}{c}{Image Modality} && \multicolumn{3}{c}{Text Modality} && \multicolumn{3}{c}{Multi-modal Fusion} && \multicolumn{1}{c}{\#Nb. Samples} \\
         \noalign{\smallskip}\hline\noalign{\smallskip}
          && Precision & Recall & F1-Score &&  Precision & Recall & F1-Score && Precision & Recall & F1-Score\\
         \cmidrule{3-14}
         Advertisement 
         && 0.9910 & 0.9565 & 0.9735
         && 0.9911 & 0.9696 & 0.9802 
         && 0.9865 & 0.9565 & 0.9713 
         && 230 \\
         Email 
         && 0.9916 & 0.99 & 0.9908 
         && 0.9933 & 0.99 & 0.9916 
         && 0.99 & 0.99 & 0.99 
         && 599 \\
         Form 
         && 0.9628 & 0.9606 & 0.9617 
         && 0.9630 & 0.9652 & 0.9641 
         && 0.9627 & 0.9582 & 0.9605 
         && 431 \\
         Letter 
         && 0.8983 & 0.9965 & 0.9448
         && 0.9040 & 0.9965 & 0.9480 
         && 0.9056 & 0.9982 & 0.9497 
         && 567 \\
         Memo 
         && 0.9857 & 1 & 0.9928 
         && 0.9841 & 1 & 0.9920 
         && 0.9857 & 1 & 0.9928 
         && 620 \\
         News article 
         && 0.9490 & 0.9894 & 0.9688
         && 0.9588 & 0.9894 & 0.9738 
         && 0.9487 & 0.9840 & 0.9661 
         && 188 \\
         Resume 
         && 0.9917 & 1 & 0.9959 
         && 0.9917 & 1 & 0.9959 
         && 0.9836 & 1 & 0.9917 
         && 120 \\
         Scientific publication 
         && 0.9519 & 0.3793 & 0.5425
         && 0.9592 & 0.3602 & 0.5237 
         && 0.9364 & 0.3946 & 0.5553 
         && 261 \\
         Scientific report 
         && 0.2374 & 0.1774 & 0.2030
         && 0.2261 & 0.1698 & 0.1940 
         && 0.2709 & 0.2075 & 0.2350 
         && 265 \\
    \noalign{\smallskip}\hline\noalign{\smallskip}
        Overall Accuracy (\%) && & & 87.29 &&  &  & 87.23 &&  &  & 87.63 \\
    \noalign{\smallskip}\hline
    \end{tabular}%
    }
    \label{tab:Evaluation of APML on tobacco}
\end{table*}


\setlength\tabcolsep{3.8 pt}
\begin{table*}[tbh]
\small
\centering
\caption{The Inter-Dataset Evaluation results of the Ensemble Self-Attention Mutual Learning (EAML$_{{Tr-KLD}_{Reg}}$) approach on the RVL-CDIP dataset.}
\resizebox{\textwidth}{!} {%
    \begin{tabular}{lcccccccccccc}
    \hline\noalign{\smallskip}
     \multicolumn{1}{c}{} && \multicolumn{10}{c}{Ensemble Self-Attention Mutual Learning (EAML$_{{Tr-KLD}_{Reg}}$)} \\
         \noalign{\smallskip}\hline\noalign{\smallskip}
         \multicolumn{1}{c}{Class Labels} && \multicolumn{3}{c}{Image Modality} && \multicolumn{3}{c}{Text Modality} && \multicolumn{3}{c}{Multi-modal Fusion}\\
         \noalign{\smallskip}\hline\noalign{\smallskip}
          && Precision & Recall & F1-Score &&  Precision & Recall & F1-Score && Precision & Recall & F1-Score\\
         \cmidrule{3-13}
         Advertisement 
         && 0.8292 & 0.9337 & 0.8783
         && 0.8702 & 0.7281 & 0.7929 
         && 0.9381 & 0.9769 & 0.9571 
         \\
         Email 
         && 0.9654 & 0.9799 & 0.9726
         && 0.9820 & 0.9366 & 0.9588 
         && 0.9944 & 0.9964 & 0.9954 
         \\
         Form 
         && 0.7953 & 0.9126 & 0.8499 
         && 0.9289 & 0.8746 & 0.9009 
         && 0.9588 & 0.9846 & 0.9715 
         \\
         Letter 
         && 0.9763 & 0.8109 & 0.8859
         && 0.9417 & 0.8816 & 0.9106 
         && 0.9970 & 0.9574 & 0.9768
         \\
         Memo 
         && 0.9660 & 0.8874 & 0.9250 
         && 0.9630 & 0.8926 & 0.9265 
         && 0.9972 & 0.9729 & 0.9849 
         \\
         News article 
         && 0.9577 & 0.7579 & 0.8462
         && 0.9574 & 0.8076 & 0.8762 
         && 0.9966 & 0.9197 & 0.9566 
         \\
         Resume 
         && 0.9811 & 0.8802 & 0.9279
         && 0.9985 & 0.9718 & 0.9850 
         && 0.9998 & 0.9891 & 0.9944 
         \\
         Scientific publication 
         && 0.5298 & 0.8827 & 0.6622
         && 0.5218 & 0.9268 & 0.6677 
         && 0.5203 & 0.9856 & 0.6810
         \\
         Scientific report 
         && 0.1858 & 0.0565 & 0.0867
         && 0.3246 & 0.0974 & 0.1498 
         && 0.2889 & 0.0197 & 0.0368 
         \\
    \noalign{\smallskip}\hline\noalign{\smallskip}
        Overall Accuracy (\%) && & & 78.89 &&  &  & 79.06 &&  &  & 86.68 \\
    \noalign{\smallskip}\hline
    
    \end{tabular}%
    }
    \label{tab:Evaluation of APML on RVLCDIP}
    
\end{table*}


\subsection{Inter-dataset Evaluation}

\subsubsection{Evaluation on the Tobacco-3482 Dataset}
\label{subsubsec:Evaluation on the Tobacco-3482 Dataset}

To evaluate the generalization ability of our ensemble network trained on the RVL-CDIP dataset, we use the benchmark Tobacco-3482 dataset and report the overall accuracy, recall, precision, and F1-score as useful metrics to evaluate the performance of the single-modal and multi-modal modalities. Since the Tobacco-3482 is an imbalanced dataset, we focus more on the precision-recall metrics which are useful to measure the success of predictions when the classes are imbalanced, which are reported in the Tables~\ref{tab:Evaluation of PML on tobacco} and \ref{tab:Evaluation of APML on tobacco}. Note that the precision metric is a measure of result relevancy, while the recall metric is a measure of how many truly relevant results are returned. The F1-score measures the weighted average of the precision and recall, while the relative contribution of precision and recall to the F1-score are equal.                                                                
However, we evaluate on 9 classes of the RVL-CDIP dataset which overlap with the classes of the Tobacco-3482 dataset, that are: Advertisement, Email, Form, Letter, Memo, News article, Resume, Scientific publication, and Scientific report. We exclude the category named Note from the Tobacco-3482 dataset which does not overlap with any of the categories of the RVL-CDIP dataset.

As it can be seen from the Tables.~(\ref{tab:Evaluation of PML on tobacco}, \ref{tab:Evaluation of APML on tobacco}) and the (Online Resource 1, Figures.(3, 4, 5)), the proposed EAML$_{{Tr-KLD}_{Reg}}$ method displays a better generalization behavior than the ML$_{{Tr-KLD}_{Reg}}$ method (Online Resource 1, Figures.(6, 7, 8)) over 8 categories that overlap with the RVL-CDIP dataset. The EAML$_{{Tr-KLD}_{Reg}}$ method performs better with an overall accuracy of $87.29\%$ for the image modality, $87.23\%$ for the text modality, and $87.63\%$ for the multi-modal fusion modality, compared to $84.82\%$, $83.72\%$, and $86.68\%$ for the ML$_{{Tr-KLD}_{Reg}}$ method respectively.
Regarding the Scientific publication category, the recall of the model considering the EAML$_{{Tr-KLD}_{Reg}}$ and ML$_{{Tr-KLD}_{Reg}}$ methods is very low. Amongst all the samples, the ability of the model to find the positive samples of the Scientific publication category is only at $37.93\%$, $36.02\%$, and $39.46\%$ for the image modality, the text modality and the multi-modal fusion modality respectively for the EAML$_{{Tr-KLD}_{Reg}}$ method, while it is at $33.72\%$, $33.72\%$, and $34.10\%$ for each modality respectively for the ML$_{{Tr-KLD}_{Reg}}$ method. The low recall for the two methods is due to the overlap between two categories that are Scientific publication and Scientific report.

After all, we see that for the two proposed EAML$_{{Tr-KLD}_{Reg}}$ and ML$_{{Tr-KLD}_{Reg}}$ methods, the model returns very few results compared to the intra-dataset evaluation, but most of its predicted labels are correct when compared to the training labels for the single-modal modalities, as well as for the multi-modal fusion modality.
Amongst all classes, the generalization ability of the model given the two methods is very poor regarding the class Scientific report, where the precision and recall are very low, whereas, for the intra-dataset evaluation, the performance of the ensemble network concerning the category Scientific report is at $94.62\%$, and $94.30\%$ for the multi-modal fusion modality of the EAML$_{{Tr-KLD}_{Reg}}$ and ML$_{{Tr-KLD}_{Reg}}$ methods respectively.
\begin{figure}[ht]
\centering
  \includegraphics[width=0.95\linewidth]{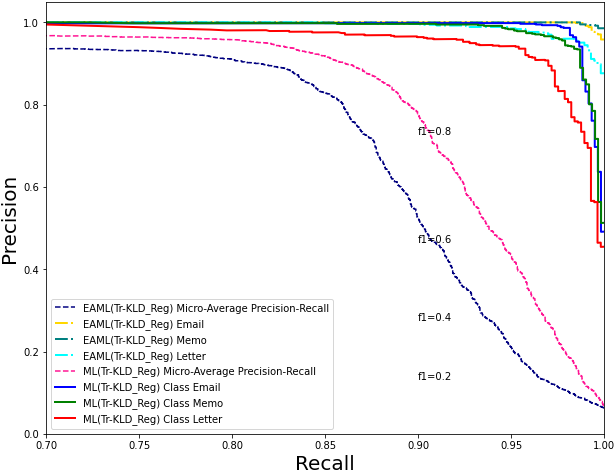}
  \caption{The Precision-Recall Curves of the Inter-Dataset Evaluation of the best classes of the Multi-Modal modalities for the two EAML$_{{Tr-KLD}_{Reg}}$ and ML$_{{Tr-KLD}_{Reg}}$ methods.}
    \label{fig:Precision_Recall_EVAL_BEST_EAML/ML}
\end{figure}

\begin{figure}[ht]
\centering
  \includegraphics[width=0.95\linewidth]{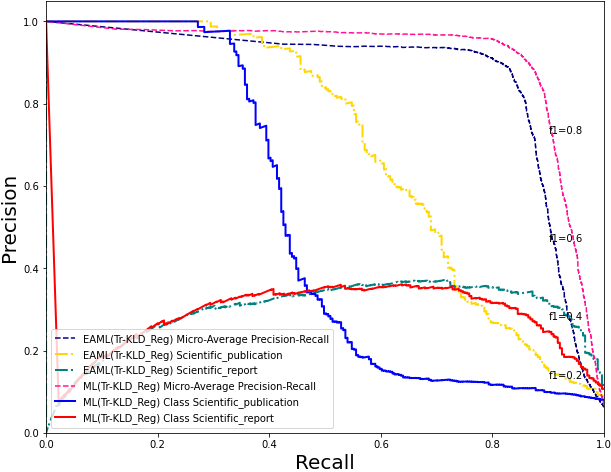}
  \caption{The Precision-Recall Curves of the Inter-Dataset Evaluation of the worst classes of the Multi-Modal modalities for the two EAML$_{{Tr-KLD}_{Reg}}$ and ML$_{{Tr-KLD}_{Reg}}$ methods.}
   \label{fig:Precision_Recall_EVAL_WORST_EAML/ML}
\end{figure}
We illustrate in the Figures.~(\ref{fig:Precision_Recall_EVAL_BEST_EAML/ML},~\ref{fig:Precision_Recall_EVAL_WORST_EAML/ML}) the precision-recall curves of the best and worst classes for the multi-modal modalities of the EAML$_{{Tr-KLD}_{Reg}}$ and ML$_{{Tr-KLD}_{Reg}}$ methods respectively. It shows the trade-off between precision and recall for different thresholds. We compute the average precision (AP) from prediction scores which summarizes a precision-recall curve. We see that the model is returning accurate results (high precision), as well as a majority of positive results (high recall), as it is the case for the categories Resume, Email, and Memo, where most of the predicted samples are labeled correctly for either the EAML$_{{Tr-KLD}_{Reg}}$ or the ML$_{{Tr-KLD}_{Reg}}$ methods. However, we observe a good precision but low recall for the Scientific publication category, and a bad precision and recall for the Scientific report category.

Therefore, Table.~\ref{tab:AP scores} illustrates the average-precision scores (AP) of the common categories for the two proposed methods ML$_{{Tr-KLD}_{Reg}}$, and EAML$_{{Tr-KLD}_{Reg}}$. Hence, we relate a good generalization ability of our proposed EAML$_{{Tr-KLD}_{Reg}}$ and ML$_{{Tr-KLD}_{Reg}}$ methods trained on RVL-CDIP, and evaluated on Tobacco-3482, regarding 7 common classes between the RVL-CDIP and Tobacco-3482 datasets, except for the Scientific publication and the Scientific report categories where it generalizes the worst. 
\setlength\tabcolsep{0.6 pt}
\begin{table}[tbh]
\small
\centering
\caption{The average precision (AP) scores of the inter-dataset evaluation of the ML$_{{Tr-KLD}_{Reg}}$ and the EAML$_{{Tr-KLD}_{Reg}}$ for the Multi-Modal Fusion modality on the Tobacco-3482 dataset}
    \begin{tabular}{lcccc}
    \hline\noalign{\smallskip}
    \multicolumn{1}{c}{} && \multicolumn{3}{c}{Method} \\
         \noalign{\smallskip}\hline\noalign{\smallskip}
     \multicolumn{1}{c}{Class Labels} && \multicolumn{1}{c}{ML$_{{Tr-KLD}_{Reg}}$} && \multicolumn{1}{c}{EAML$_{{Tr-KLD}_{Reg}}$} \\
         \noalign{\smallskip}\hline\noalign{\smallskip}
         Advertisement && 0.94 && 1.00\\
         Email && 0.99 && 1.00\\
         Form && 0.97 && 0.99\\
         Letter && 0.98 && 0.99\\
         Memo && 0.99 && 1.00\\
         News article && 0.96 && 1.00\\
         Resume && 1.00 && 1.00\\
         Scientific publication && 0.50 && 0.69\\
         Scientific report && 0.28 && 0.29\\
    \noalign{\smallskip}\hline\noalign{\smallskip}
        Micro-Average Precision && 0.86 && 0.91 \\
    \noalign{\smallskip}\hline
    
    \end{tabular}%
    \label{tab:AP scores}
    
\end{table}
\subsubsection{Evaluation on the RVL-CDIP Dataset}

Symmetrically, we propose to evaluate the generalization ability of our proposed model trained on the Tobacco-3482 dataset and validated on the large-scale RVL-CDIP dataset. The overall accuracy, recall, precision, and F1-score metrics of our best EAML$_{{Tr-KLD}_{Reg}}$ approach are proposed in the Table.~\ref{tab:Evaluation of APML on RVLCDIP}.        
We proceed with the same evaluation protocol as in Section.~\ref{subsubsec:Evaluation on the Tobacco-3482 Dataset}, where there is 9 classes of the Tobacco-3482 dataset that overlap with the classes of the RVL-CDIP dataset.

From the Table.~\ref{tab:Evaluation of APML on RVLCDIP}, the EAML$_{{Tr-KLD}_{Reg}}$ method displays a better generalization ability compared to the other methods. It performs the best with an overall accuracy of $78.89\%$ for the image modality, $79.06\%$ for the text modality, and $86.68\%$ for the multi-modal fusion modality. Amongst all classes, and similarly to the inter-dataset evaluation on the Tobacco-3482 dataset, the network generalizes the worst for the same categories which are Scientific publication and Scientific report, while it generalizes the best for the categories Resume, Letter, Memo, and Email. Moreover, the ensemble network manages to predict only $10.50\%$ of samples that belong to the Scientific report category as true positives, while $85.26\%$ are predicted as they belong to the Scientific publication category. At this stage, the precision and recall of the model are very low regarding the Scientific report category for each modality. As mentioned in section~\ref{subsubsec:Evaluation on the Tobacco-3482 Dataset}, the bad precision and recall are due to the overlap between the two categories, which results to a bad generalization ability of the EAML$_{{Tr-KLD}_{Reg}}$ method considering only the two categories, contrary to the intra-dataset evaluation, where the ensemble network achieves accurate results with high precision and recall for all the categories.

Therefore, we relate a good generalization ability of our proposed EAML$_{{Tr-KLD}_{Reg}}$ trained on Tobacco-3482, and evaluated on RVL-CDIP, regarding 7 common classes between the RVL-CDIP and Tobacco-3482 datasets, except for the Scientific publication and the Scientific report categories where it generalizes the worst. These results are encouraging as we can see that our proposed system is able to learn on a small dataset (around 6000 documents) compared to the RVL-CDIP training set.

\section{Conclusion and Future Work}
\label{sec:Conclusion and Future Work}

In this paper, we have proposed an ensemble network that jointly learns the visual structural properties and the corresponding text embeddings from document images through a self-attention-based mutual learning strategy (EAML$_{{Tr-KLD}_{Reg}}$). We have shown that the designed self-attention-based fusion module along with the mutual learning approach with the regularization term enables the current modality to learn the positive knowledge from the other modality instead of the negative knowledge, which weakens the learning capacity for the current modality during the training stage. This constraint has been realized by adding a mimicry truncated-Kullback–Leibler divergence regularization loss (\ie Tr-KLD$_{Reg}$) to the conventional supervised setting. With this approach, we have further combined the mutual predictions computed by the trained image and text modalities in an ensemble network through multi-modal learning to boost the overall classification accuracy of document images. \textcolor{black}{The proposed mutual learning strategy with regularization has shown to be efficient in improving the overall performance of the ensemble model}. For the future research, we will improve the performance and the generalization ability of our self-attention-based mutual learning strategy to enhance the learning process between different modalities both independently, and in an ensemble network.



%
%

\bibliographystyle{spmpsci}      

%
%
\bibliography{egbib.bib}

\begin{thebibliography}{10}
\providecommand{\url}[1]{{#1}}
\providecommand{\urlprefix}{URL }
\expandafter\ifx\csname urlstyle\endcsname\relax
  \providecommand{\doi}[1]{DOI~\discretionary{}{}{}#1}\else
  \providecommand{\doi}{DOI~\discretionary{}{}{}\begingroup
  \urlstyle{rm}\Url}\fi

\bibitem{Afzal2015DeepdocclassifierDC}
Afzal, M., Capobianco, S., Malik, M., Marinai, S., Breuel, T., Dengel, A.,
  Liwicki, M.: Deepdocclassifier: Document classification with deep
  convolutional neural network.
\newblock 2015 13th International Conference on Document Analysis and
  Recognition (ICDAR) pp. 1111--1115 (2015)

\bibitem{Afzal2017CuttingTE}
Afzal, M., K{\"o}lsch, A., Ahmed, S., Liwicki, M.: Cutting the error by half:
  Investigation of very deep cnn and advanced training strategies for document
  image classification.
\newblock 2017 14th IAPR International Conference on Document Analysis and
  Recognition (ICDAR) \textbf{01}, 883--888 (2017)

\bibitem{Afzal2015DocumentIB}
Afzal, M., Pastor-Pellicer, J., Shafait, F., Breuel, T., Dengel, A., Liwicki,
  M.: Document image binarization using lstm: A sequence learning approach.
\newblock In: HIP '15 (2015)

\bibitem{8578734}
{Anderson}, P., {He}, X., {Buehler}, C., {Teney}, D., {Johnson}, M., {Gould},
  S., {Zhang}, L.: Bottom-up and top-down attention for image captioning and
  visual question answering.
\newblock In: 2018 IEEE/CVF Conference on Computer Vision and Pattern
  Recognition, pp. 6077--6086 (2018).
\newblock \doi{10.1109/CVPR.2018.00636}

\bibitem{Appiani2001AutomaticDC}
Appiani, E., Cesarini, F., Colla, A., Diligenti, M., Gori, M., Marinai, S.,
  Soda, G.: Automatic document classification and indexing in high-volume
  applications.
\newblock International Journal on Document Analysis and Recognition
  \textbf{4}, 69--83 (2001)

\bibitem{Asim2019TwoSD}
Asim, M., Khan, M.U.G., Malik, M., Razzaque, K., Dengel, A., Ahmed, S.: Two
  stream deep network for document image classification.
\newblock 2019 International Conference on Document Analysis and Recognition
  (ICDAR) pp. 1410--1416 (2019)

\bibitem{audebert2019multimodal}
Audebert, N., Herold, C., Slimani, K., Vidal, C.: Multimodal deep networks for
  text and image-based document classification.
\newblock In: Joint European Conference on Machine Learning and Knowledge
  Discovery in Databases, pp. 427--443. Springer (2019)

\bibitem{Augereau2014ImprovingCO}
Augereau, O., Journet, N., Vialard, A., Domenger, J.P.: Improving
  classification of an industrial document image database by combining visual
  and textual features.
\newblock 2014 11th IAPR International Workshop on Document Analysis Systems
  pp. 314--318 (2014)

\bibitem{Bahdanau2015NeuralMT}
Bahdanau, D., Cho, K., Bengio, Y.: Neural machine translation by jointly
  learning to align and translate.
\newblock CoRR \textbf{abs/1409.0473} (2015)

\bibitem{9191268}
{Bakkali}, S., {Ming}, Z., {Coustaty}, M., {Rusiñol}, M.: Cross-modal deep
  networks for document image classification.
\newblock In: 2020 IEEE International Conference on Image Processing (ICIP),
  pp. 2556--2560 (2020).
\newblock \doi{10.1109/ICIP40778.2020.9191268}

\bibitem{souhailbakkali}
{Bakkali}, S., {Ming}, Z., {Coustaty}, M., {Rusiñol}, M.: Visual and textual
  deep feature fusion for document image classification.
\newblock In: 2020 IEEE/CVF Conference on Computer Vision and Pattern
  Recognition Workshops (CVPRW), pp. 2394--2403 (2020).
\newblock \doi{10.1109/CVPRW50498.2020.00289}

\bibitem{Byun2000FormCU}
Byun, Y., Lee, Y.: Form classification using dp matching.
\newblock In: SAC '00 (2000)

\bibitem{chen2016abccnn}
Chen, K., Wang, J., Chen, L.C., Gao, H., Xu, W., Nevatia, R.: Abc-cnn: An
  attention based convolutional neural network for visual question answering
  (2016)

\bibitem{Chen2006ASO}
Chen, N., Blostein, D.: A survey of document image classification: problem
  statement, classifier architecture and performance evaluation.
\newblock International Journal of Document Analysis and Recognition (IJDAR)
  \textbf{10}, 1--16 (2006)

\bibitem{Csurka2016WhatIT}
Csurka, G., Larlus, D., Gordo, A., Almaz{\'a}n, J.: What is the right way to
  represent document images?
\newblock ArXiv \textbf{abs/1603.01076} (2016)

\bibitem{8545630}
{Das}, A., {Roy}, S., {Bhattacharya}, U., {Parui}, S.K.: Document image
  classification with intra-domain transfer learning and stacked generalization
  of deep convolutional neural networks.
\newblock In: 2018 24th International Conference on Pattern Recognition (ICPR),
  pp. 3180--3185 (2018).
\newblock \doi{10.1109/ICPR.2018.8545630}

\bibitem{Dauphinee2019ModularMA}
Dauphinee, T., Patel, N., Rashidi, M.M.: Modular multimodal architecture for
  document classification.
\newblock ArXiv \textbf{abs/1912.04376} (2019)

\bibitem{Dengel1995ClusteringAC}
Dengel, A., Dubiel, F.: Clustering and classification of document structure-a
  machine learning approach.
\newblock Proceedings of 3rd International Conference on Document Analysis and
  Recognition \textbf{2}, 587--591 vol.2 (1995)

\bibitem{Devlin2019BERTPO}
Devlin, J., Chang, M.W., Lee, K., Toutanova, K.: Bert: Pre-training of deep
  bidirectional transformers for language understanding.
\newblock In: NAACL-HLT (2019)

\bibitem{devries2017improved}
DeVries, T., Taylor, G.W.: Improved regularization of convolutional neural
  networks with cutout (2017)

\bibitem{Ferrando2020ImprovingAA}
Ferrando, J., Dom{\'i}nguez, J.L., Torres, J., Garc{\'i}a, R., Garc{\'i}a, D.,
  Garrido, D., Cortada, J., Valero, M.: Improving accuracy and speeding up
  document image classification through parallel systems.
\newblock Computational Science – ICCS 2020 \textbf{12138}, 387 -- 400 (2020)

\bibitem{Fukui2016MultimodalCB}
Fukui, A., Park, D.H., Yang, D., Rohrbach, A., Darrell, T., Rohrbach, M.:
  Multimodal compact bilinear pooling for visual question answering and visual
  grounding.
\newblock ArXiv \textbf{abs/1606.01847} (2016)

\bibitem{Gallo2018ImageAE}
Gallo, I., Calefati, A., Nawaz, S., Janjua, M.K.: Image and encoded text fusion
  for multi-modal classification.
\newblock 2018 Digital Image Computing: Techniques and Applications (DICTA) pp.
  1--7 (2018)

\bibitem{Hao2016ATD}
Hao, L., Gao, L., Yi, X., Tang, Z.: A table detection method for pdf documents
  based on convolutional neural networks.
\newblock 2016 12th IAPR Workshop on Document Analysis Systems (DAS) pp.
  287--292 (2016)

\bibitem{Harley2015EvaluationOD}
Harley, A.W., Ufkes, A., Derpanis, K.: Evaluation of deep convolutional nets
  for document image classification and retrieval.
\newblock 2015 13th International Conference on Document Analysis and
  Recognition (ICDAR) pp. 991--995 (2015)

\bibitem{He2016DeepRL}
He, K., Zhang, X., Ren, S., Sun, J.: Deep residual learning for image
  recognition.
\newblock 2016 IEEE Conference on Computer Vision and Pattern Recognition
  (CVPR) pp. 770--778 (2016)

\bibitem{hinton2015distilling}
Hinton, G., Vinyals, O., Dean, J.: Distilling the knowledge in a neural
  network.
\newblock arXiv preprint arXiv:1503.02531  (2015)

\bibitem{Hu_2018_CVPR}
Hu, J., Shen, L., Sun, G.: Squeeze-and-excitation networks.
\newblock In: Proceedings of the IEEE Conference on Computer Vision and Pattern
  Recognition (CVPR) (2018)

\bibitem{Kang2014ConvolutionalNN}
Kang, L., Kumar, J., Ye, P., Li, Y., Doermann, D.: Convolutional neural
  networks for document image classification.
\newblock 2014 22nd International Conference on Pattern Recognition pp.
  3168--3172 (2014)

\bibitem{6977258}
{Kang}, L., {Kumar}, J., {Ye}, P., {Li}, Y., {Doermann}, D.: Convolutional
  neural networks for document image classification.
\newblock In: 2014 22nd International Conference on Pattern Recognition, pp.
  3168--3172 (2014).
\newblock \doi{10.1109/ICPR.2014.546}

\bibitem{NEURIPS2018_96ea64f3}
Kim, J.H., Jun, J., Zhang, B.T.: Bilinear attention networks.
\newblock In: S.~Bengio, H.~Wallach, H.~Larochelle, K.~Grauman,
  N.~Cesa-Bianchi, R.~Garnett (eds.) Advances in Neural Information Processing
  Systems, vol.~31, pp. 1564--1574. Curran Associates, Inc. (2018).
\newblock
  \urlprefix\url{https://proceedings.neurips.cc/paper/2018/file/96ea64f3a1aa2fd00c72faacf0cb8ac9-Paper.pdf}

\bibitem{Klsch2017RealTimeDI}
K{\"o}lsch, A., Afzal, M., Ebbecke, M., Liwicki, M.: Real-time document image
  classification using deep cnn and extreme learning machines.
\newblock 2017 14th IAPR International Conference on Document Analysis and
  Recognition (ICDAR) \textbf{01}, 1318--1323 (2017)

\bibitem{Krizhevsky2017ImageNetCW}
Krizhevsky, A., Sutskever, I., Hinton, G.E.: Imagenet classification with deep
  convolutional neural networks.
\newblock In: CACM (2017)

\bibitem{Kumar2014StructuralSF}
Kumar, J., Ye, P., Doermann, D.: Structural similarity for document image
  classification and retrieval.
\newblock Pattern Recognit. Lett. \textbf{43}, 119--126 (2014)

\bibitem{Lai2015RecurrentCN}
Lai, S., Xu, L., Liu, K., Zhao, J.: Recurrent convolutional neural networks for
  text classification.
\newblock In: AAAI (2015)

\bibitem{LeCun1998GradientbasedLA}
{Lecun}, Y., {Bottou}, L., {Bengio}, Y., {Haffner}, P.: Gradient-based learning
  applied to document recognition.
\newblock Proceedings of the IEEE \textbf{86}(11), 2278--2324 (1998).
\newblock \doi{10.1109/5.726791}

\bibitem{Lee2018StackedCA}
Lee, K.H., Chen, X., Hua, G., Hu, H., He, X.: Stacked cross attention for
  image-text matching.
\newblock In: ECCV (2018)

\bibitem{Li2019VisualSR}
Li, K., Zhang, Y., Li, K., Li, Y., Fu, Y.: Visual semantic reasoning for
  image-text matching.
\newblock 2019 IEEE/CVF International Conference on Computer Vision (ICCV) pp.
  4653--4661 (2019)

\bibitem{lu2017hierarchical}
Lu, J., Yang, J., Batra, D., Parikh, D.: Hierarchical question-image
  co-attention for visual question answering.
\newblock ArXiv \textbf{abs/1606.00061} (2016)

\bibitem{Mikolov2013EfficientEO}
Mikolov, T., Chen, K., Corrado, G.S., Dean, J.: Efficient estimation of word
  representations in vector space.
\newblock CoRR \textbf{abs/1301.3781} (2013)

\bibitem{Mikolov2018AdvancesIP}
Mikolov, T., Grave, E., Bojanowski, P., Puhrsch, C., Joulin, A.: Advances in
  pre-training distributed word representations.
\newblock ArXiv \textbf{abs/1712.09405} (2018)

\bibitem{Nguyen2018ImprovedFO}
Nguyen, D.K., Okatani, T.: Improved fusion of visual and language
  representations by dense symmetric co-attention for visual question
  answering.
\newblock 2018 IEEE/CVF Conference on Computer Vision and Pattern Recognition
  pp. 6087--6096 (2018)

\bibitem{Noce2016EmbeddedTC}
Noce, L., Gallo, I., Zamberletti, A., Calefati, A.: Embedded textual content
  for document image classification with convolutional neural networks.
\newblock In: DocEng '16 (2016)

\bibitem{PastorPellicer2016CompleteSF}
Pastor-Pellicer, J., Afzal, M., Liwicki, M., Bleda, M.J.: Complete system for
  text line extraction using convolutional neural networks and watershed
  transform.
\newblock 2016 12th IAPR Workshop on Document Analysis Systems (DAS) pp. 30--35
  (2016)

\bibitem{PastorPellicer2015InsightsOT}
Pastor-Pellicer, J., Boquera, S.E., Zamora-Mart{\'i}nez, F., Afzal, M.Z.,
  Bleda, M.J.C.: Insights on the use of convolutional neural networks for
  document image binarization.
\newblock In: IWANN (2015)

\bibitem{Pennington2014GloveGV}
Pennington, J., Socher, R., Manning, C.D.: Glove: Global vectors for word
  representation.
\newblock In: EMNLP (2014)

\bibitem{Peters2018DeepCW}
Peters, M.E., Neumann, M., Iyyer, M., Gardner, M., Clark, C., Lee, K.,
  Zettlemoyer, L.: Deep contextualized word representations.
\newblock ArXiv \textbf{abs/1802.05365} (2018)

\bibitem{Qian2010ANA}
Qian, J., Wang, W., Wang, D.: A novel approach for online handwriting
  recognition of tibetan characters (2010)

\bibitem{ramachandran2019standalone}
Ramachandran, P., Parmar, N., Vaswani, A., Bello, I., Levskaya, A., Shlens, J.:
  Stand-alone self-attention in vision models  (2019)

\bibitem{Russakovsky2015ImageNetLS}
Russakovsky, O., Deng, J., Su, H., Krause, J., Satheesh, S., Ma, S., Huang, Z.,
  Karpathy, A., Khosla, A., Bernstein, M.S., Berg, A., Fei-Fei, L.: Imagenet
  large scale visual recognition challenge.
\newblock International Journal of Computer Vision \textbf{115}, 211--252
  (2015)

\bibitem{Seuret2017PCAInitializedDN}
Seuret, M., Alberti, M., Liwicki, M., Ingold, R.: Pca-initialized deep neural
  networks applied to document image analysis.
\newblock 2017 14th IAPR International Conference on Document Analysis and
  Recognition (ICDAR) \textbf{01}, 877--882 (2017)

\bibitem{Sierra2018CombiningTA}
Sierra, S., Gonz{\'a}lez, F.A.: Combining textual and visual representations
  for multimodal author profiling: Notebook for pan at clef 2018.
\newblock In: CLEF (2018)

\bibitem{Simonyan2015VeryDC}
Simonyan, K., Zisserman, A.: Very deep convolutional networks for large-scale
  image recognition.
\newblock CoRR \textbf{abs/1409.1556} (2015)

\bibitem{Szegedy2017Inceptionv4IA}
Szegedy, C., Ioffe, S., Vanhoucke, V., Alemi, A.A.: Inception-v4,
  inception-resnet and the impact of residual connections on learning.
\newblock In: AAAI (2017)

\bibitem{Szegedy2015GoingDW}
Szegedy, C., Liu, W., Jia, Y., Sermanet, P., Reed, S., Anguelov, D., Erhan, D.,
  Vanhoucke, V., Rabinovich, A.: Going deeper with convolutions.
\newblock 2015 IEEE Conference on Computer Vision and Pattern Recognition
  (CVPR) pp. 1--9 (2015)

\bibitem{tensmeyer2017analysis}
{Tensmeyer}, C., {Martinez}, T.: Analysis of convolutional neural networks for
  document image classification.
\newblock In: 2017 14th IAPR International Conference on Document Analysis and
  Recognition (ICDAR), vol.~01, pp. 388--393 (2017).
\newblock \doi{10.1109/ICDAR.2017.71}

\bibitem{UlHasan2015ASL}
Ul-Hasan, A., Afzal, M., Shafait, F., Liwicki, M., Breuel, T.: A sequence
  learning approach for multiple script identification.
\newblock 2015 13th International Conference on Document Analysis and
  Recognition (ICDAR) pp. 1046--1050 (2015)

\bibitem{vaswani2017attention}
Vaswani, A., Shazeer, N., Parmar, N., Uszkoreit, J., Jones, L., Gomez, A.N.,
  Kaiser, L., Polosukhin, I.: Attention is all you need.
\newblock ArXiv \textbf{abs/1706.03762} (2017)

\bibitem{Wang2018NonlocalNN}
Wang, X., Girshick, R.B., Gupta, A., He, K.: Non-local neural networks.
\newblock 2018 IEEE/CVF Conference on Computer Vision and Pattern Recognition
  pp. 7794--7803 (2018)

\bibitem{wang2019position}
Wang, Y., Yang, H., Qian, X., Ma, L., Lu, J., Li, B., Fan, X.: Position focused
  attention network for image-text matching.
\newblock In: Proceedings of the Twenty-Eighth International Joint Conference
  on Artificial Intelligence, {IJCAI-19}, pp. 3792--3798. International Joint
  Conferences on Artificial Intelligence Organization (2019).
\newblock \doi{10.24963/ijcai.2019/526}.
\newblock \urlprefix\url{https://doi.org/10.24963/ijcai.2019/526}

\bibitem{Xu2020LayoutLMPO}
Xu, Y., Li, M., Cui, L., Huang, S., Wei, F., Zhou, M.: Layoutlm: Pre-training
  of text and layout for document image understanding.
\newblock Proceedings of the 26th ACM SIGKDD International Conference on
  Knowledge Discovery \& Data Mining  (2020)

\bibitem{Xu2020LayoutLMv2MP}
Xu, Y., Xu, Y., Lv, T., Cui, L., Wei, F., Wang, G., Lu, Y., Flor{\^e}ncio, D.,
  Zhang, C., Che, W., Zhang, M., Zhou, L.: Layoutlmv2: Multi-modal pre-training
  for visually-rich document understanding.
\newblock ArXiv \textbf{abs/2012.14740} (2020)

\bibitem{Yan2020ImageCV}
Yan, S., Xie, Y., Wu, F., Smith, J., Lu, W., Zhang, B.: Image captioning via
  hierarchical attention mechanism and policy gradient optimization.
\newblock Signal Process. \textbf{167} (2020)

\bibitem{Yang2019ExploringDM}
Yang, F., Peng, X., Ghosh, G., Shilon, R., Ma, H., Moore, E., Predovic, G.:
  Exploring deep multimodal fusion of text and photo for hate speech
  classification.
\newblock In: Proceedings of the Third Workshop on Abusive Language Online, pp.
  11--18. Association for Computational Linguistics, Florence, Italy (2019).
\newblock \doi{10.18653/v1/W19-3502}.
\newblock \urlprefix\url{https://www.aclweb.org/anthology/W19-3502}

\bibitem{Yang2017LearningTE}
Yang, X., Yumer, E., Asente, P., Kraley, M., Kifer, D., Giles, C.L.: Learning
  to extract semantic structure from documents using multimodal fully
  convolutional neural networks.
\newblock 2017 IEEE Conference on Computer Vision and Pattern Recognition
  (CVPR) pp. 4342--4351 (2017)

\bibitem{Yang2019XLNetGA}
Yang, Z., Dai, Z., Yang, Y., Carbonell, J., Salakhutdinov, R., Le, Q.V.: Xlnet:
  Generalized autoregressive pretraining for language understanding.
\newblock In: NeurIPS (2019)

\bibitem{Yang2016StackedAN}
Yang, Z., He, X., Gao, J., Deng, L., Smola, A.: Stacked attention networks for
  image question answering.
\newblock 2016 IEEE Conference on Computer Vision and Pattern Recognition
  (CVPR) pp. 21--29 (2016)

\bibitem{Yu2019MultimodalUA}
Yu, Z., Cui, Y., Yu, J., Tao, D., Tian, Q.: Multimodal unified attention
  networks for vision-and-language interactions.
\newblock ArXiv \textbf{abs/1908.04107} (2019)

\bibitem{Yu_2017_ICCV}
Yu, Z., Yu, J., Fan, J., Tao, D.: Multi-modal factorized bilinear pooling with
  co-attention learning for visual question answering.
\newblock In: Proceedings of the IEEE International Conference on Computer
  Vision (ICCV) (2017)

\bibitem{Yu_2018}
{Yu}, Z., {Yu}, J., {Xiang}, C., {Fan}, J., {Tao}, D.: Beyond bilinear:
  Generalized multimodal factorized high-order pooling for visual question
  answering.
\newblock IEEE Transactions on Neural Networks and Learning Systems
  \textbf{29}(12), 5947--5959 (2018).
\newblock \doi{10.1109/TNNLS.2018.2817340}

\bibitem{zahavy2016picture}
Zahavy, T., Magnani, A., Krishnan, A., Mannor, S.: Is a picture worth a
  thousand words? a deep multi-modal fusion architecture for product
  classification in e-commerce (2016)

\bibitem{zhang2018deep}
Zhang, Y., Xiang, T., Hospedales, T.M., Lu, H.: Deep mutual learning.
\newblock In: Proceedings of the IEEE Conference on Computer Vision and Pattern
  Recognition, pp. 4320--4328 (2018)

\bibitem{Zhao2020ExploringSF}
Zhao, H., Jia, J., Koltun, V.: Exploring self-attention for image recognition.
\newblock 2020 IEEE/CVF Conference on Computer Vision and Pattern Recognition
  (CVPR) pp. 10073--10082 (2020)

\bibitem{Zhou2015SimpleBF}
Zhou, B., Tian, Y., Sukhbaatar, S., Szlam, A., Fergus, R.: Simple baseline for
  visual question answering.
\newblock ArXiv \textbf{abs/1512.02167} (2015)

\end{thebibliography}

\end{document}


\title{Supplementary Material of IJDAR-ICDAR 2021 Journal Track Paper
}

\titlerunning{EAML}        

\author{Souhail Bakkali         \and
        Zuheng Ming             \and 
        Mickaël Coustaty        \and 
        Marçal Rusiñol          
}

\authorrunning{Souhail Bakkali et al.}

\institute{Souhail Bakkali \at
           L3i, University of La Rochelle, La Rochelle, France \\
           \email{souhail.bakkali@univ-lr.fr}      
           \and
           Zuheng Ming \at
           L3i, University of La Rochelle, La Rochelle, France \\
           \email{zuheng.ming@univ-lr.fr}          
           \and
           Mickaël Coustaty \at
           L3i, University of La Rochelle, La Rochelle, France \\
           \email{mickael.coustaty@univ-lr.fr}          
           \and
           Marçal Rusiñol \at
           AllRead Machine Learning Technologies, Barcelona, Spain\\
           \email{marcal@allread.ai}           
           \and
}

\date{Received: date / Accepted: date}

\maketitle

\section{Confusion Matrices}
\subsection{Intra-Dataset Confusion Matrices}
\label{Supplementary}
\begin{figure*}[ht]
\centering
  \includegraphics[width=0.67\linewidth]{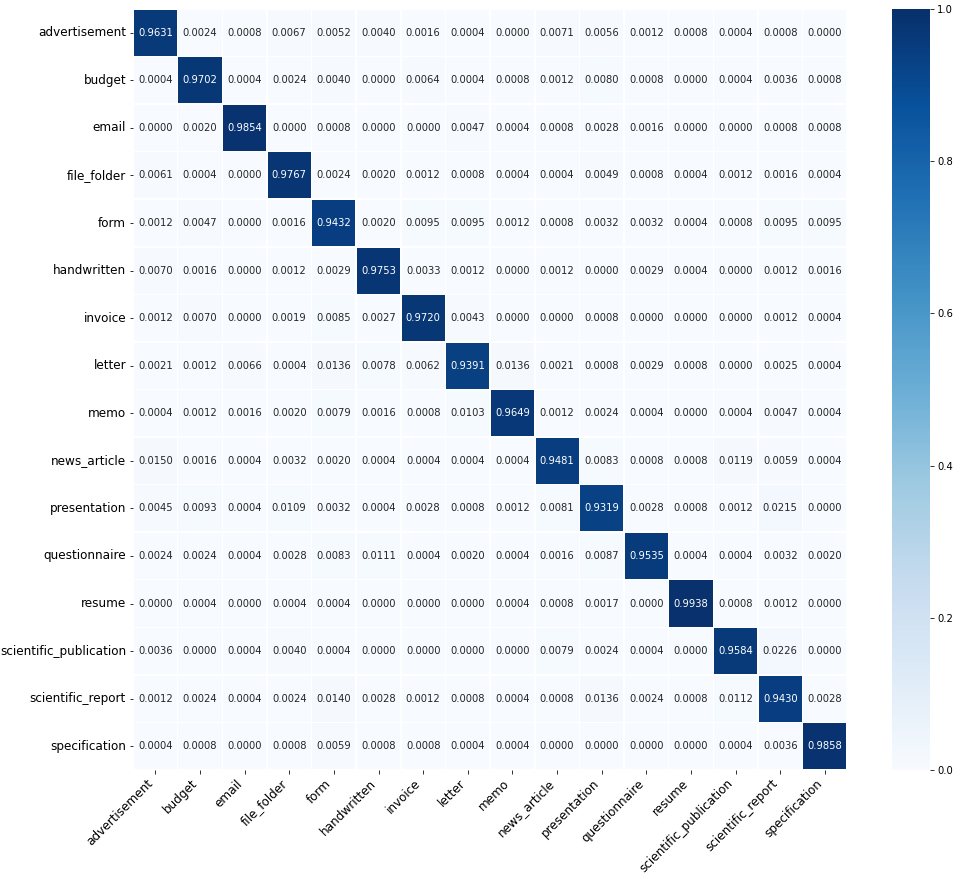}
  \caption{Multi-Modal Fusion Modality of the ML$_{{Tr-KLD}_{Reg}}$ method.}
    \label{fig:Multi-Modal_ML}
\end{figure*}
\begin{figure*}[ht]
\centering
  \includegraphics[width=0.67\linewidth]{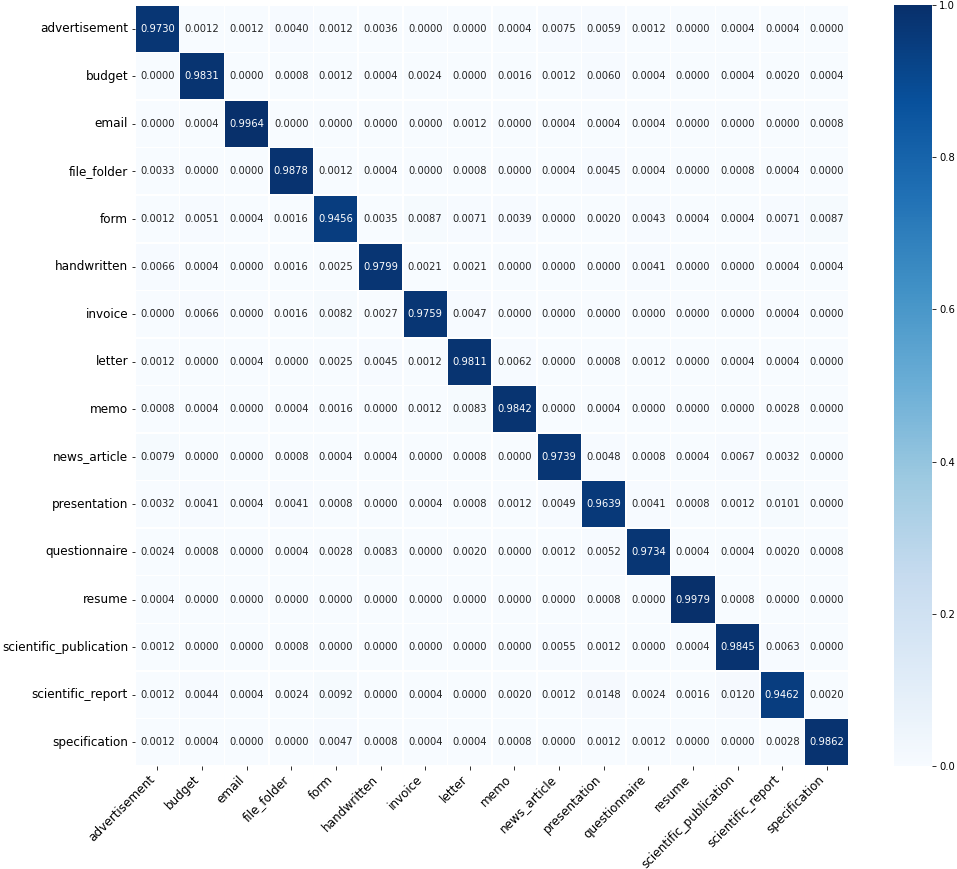}
  \caption{Multi-Modal Fusion Modality of the EAML$_{{Tr-KLD}_{Reg}}$ method.}
    \label{fig:Multi-Modal_EAML}
\end{figure*}

The Figures.~(\ref{fig:Multi-Modal_ML},~\ref{fig:Multi-Modal_EAML}) illustrate the confusion matrices of the EAML$_{{Tr-KLD}_{Reg}}$ and ML$_{{Tr-KLD}_{Reg}}$ methods. The figures show that the combined predictions from the image and text modalities through a fusion methodology improve the classification accuracy of each class of the dataset independently, compared to the single image and text modalities. Furthermore, the EAML$_{{Tr-KLD}_{Reg}}$ method outperforms the ML$_{{Tr-KLD}_{Reg}}$ methods given the multi-modal fusion classification results.

\subsection{Inter-Dataset Confusion Matrices}
\begin{figure}[ht]
\centering
  \includegraphics[width=1\linewidth]{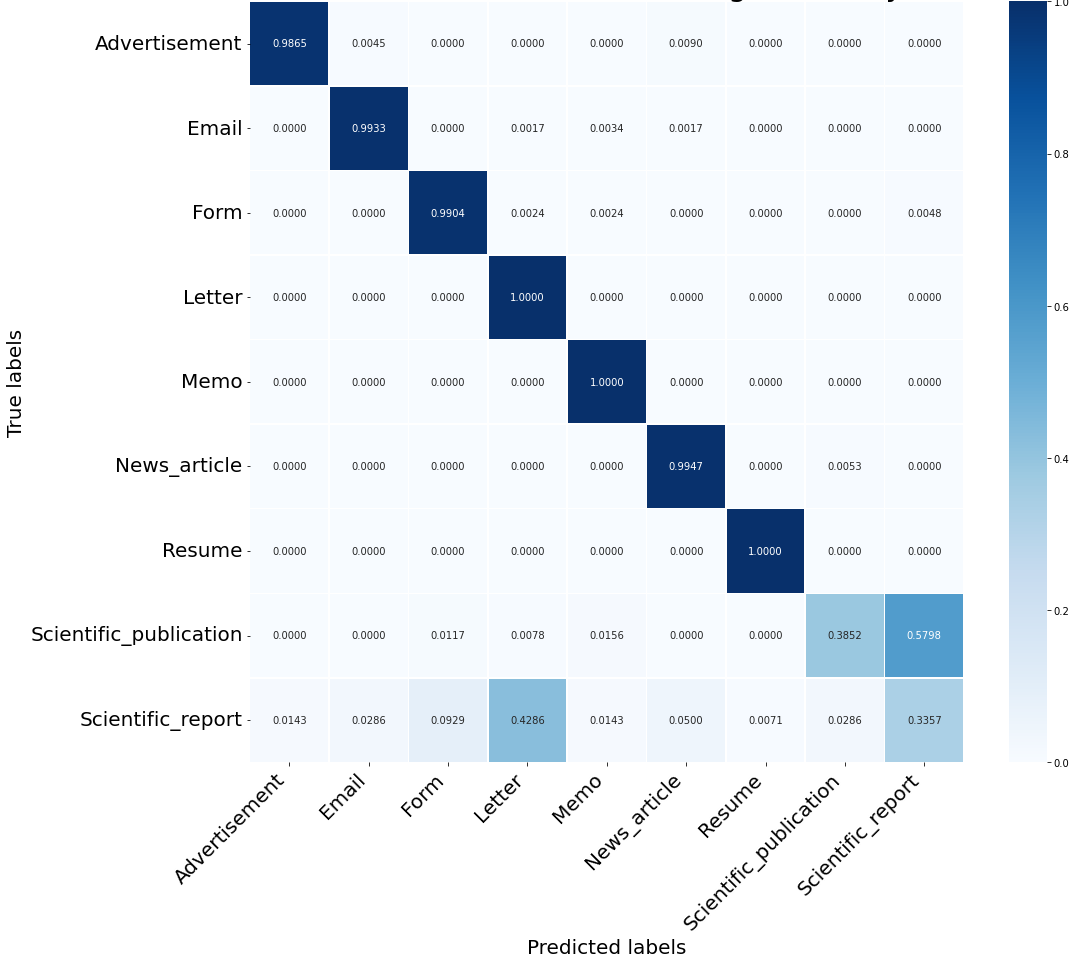}
  \caption{The Image Modality Confusion Matrix of the EAML$_{{Tr-KLD}_{Reg}}$) method.}
    \label{fig:CM_EVAL_Image_APML}
\end{figure}

\begin{figure}[ht]
\centering
  \includegraphics[width=1\linewidth]{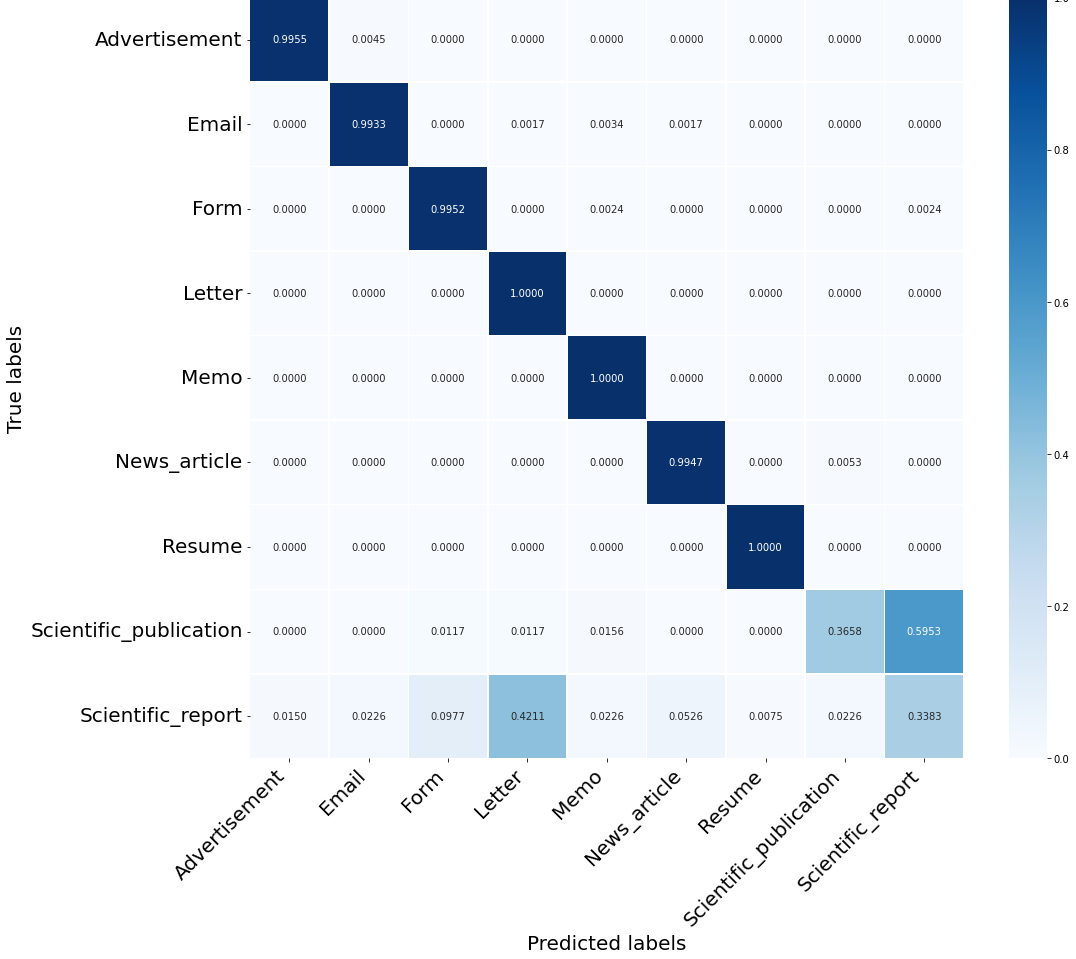}
  \caption{The Text Modality Confusion Matrix of the EAML$_{{Tr-KLD}_{Reg}}$ method.}
    \label{fig:CM_EVAL_Text_APML}
\end{figure}

\begin{figure}[ht]
\centering
  \includegraphics[width=1\linewidth]{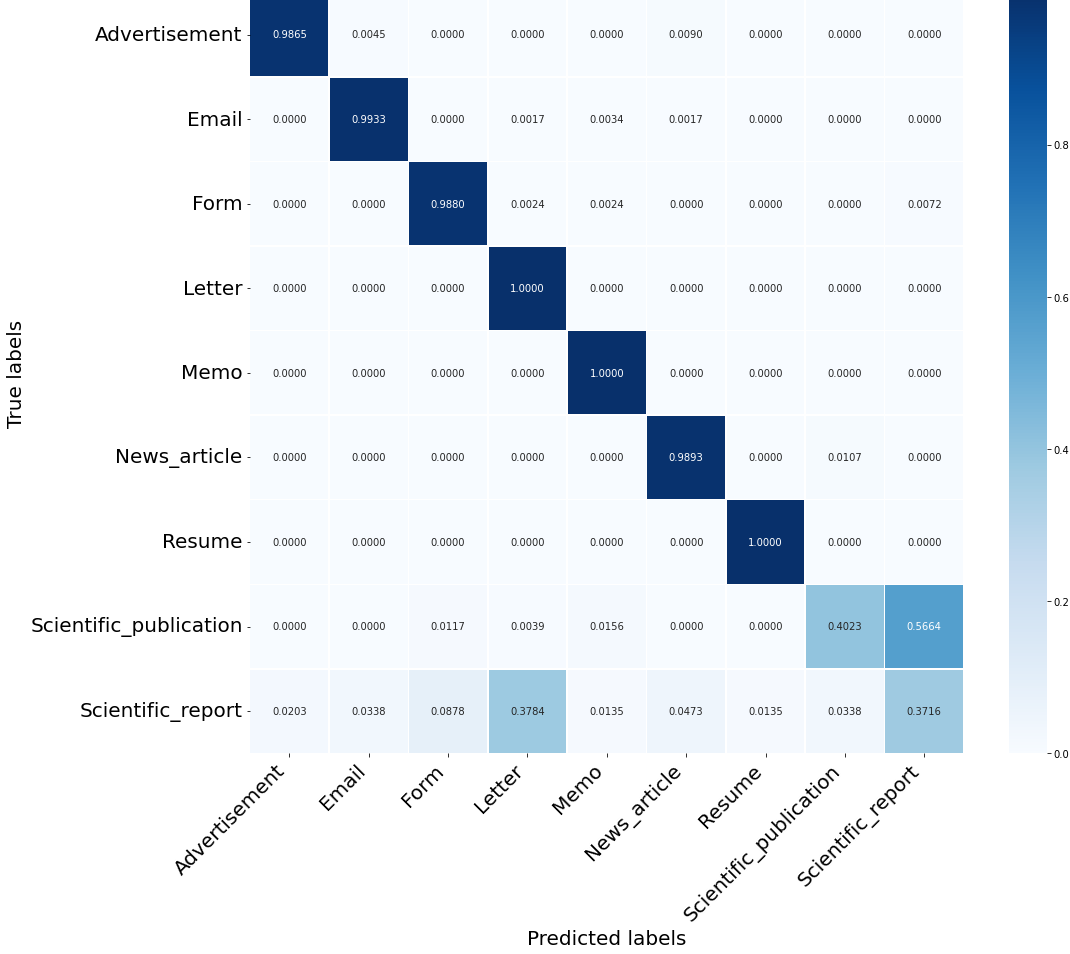}
  \caption{The Multi-Modal Fusion Modality Confusion Matrix of the EAML$_{{Tr-KLD}_{Reg}}$ method.}
    \label{fig:CM_EVAL_MultiModal_APML}
\end{figure}

\begin{figure}[ht]
\centering
  \includegraphics[width=1\linewidth]{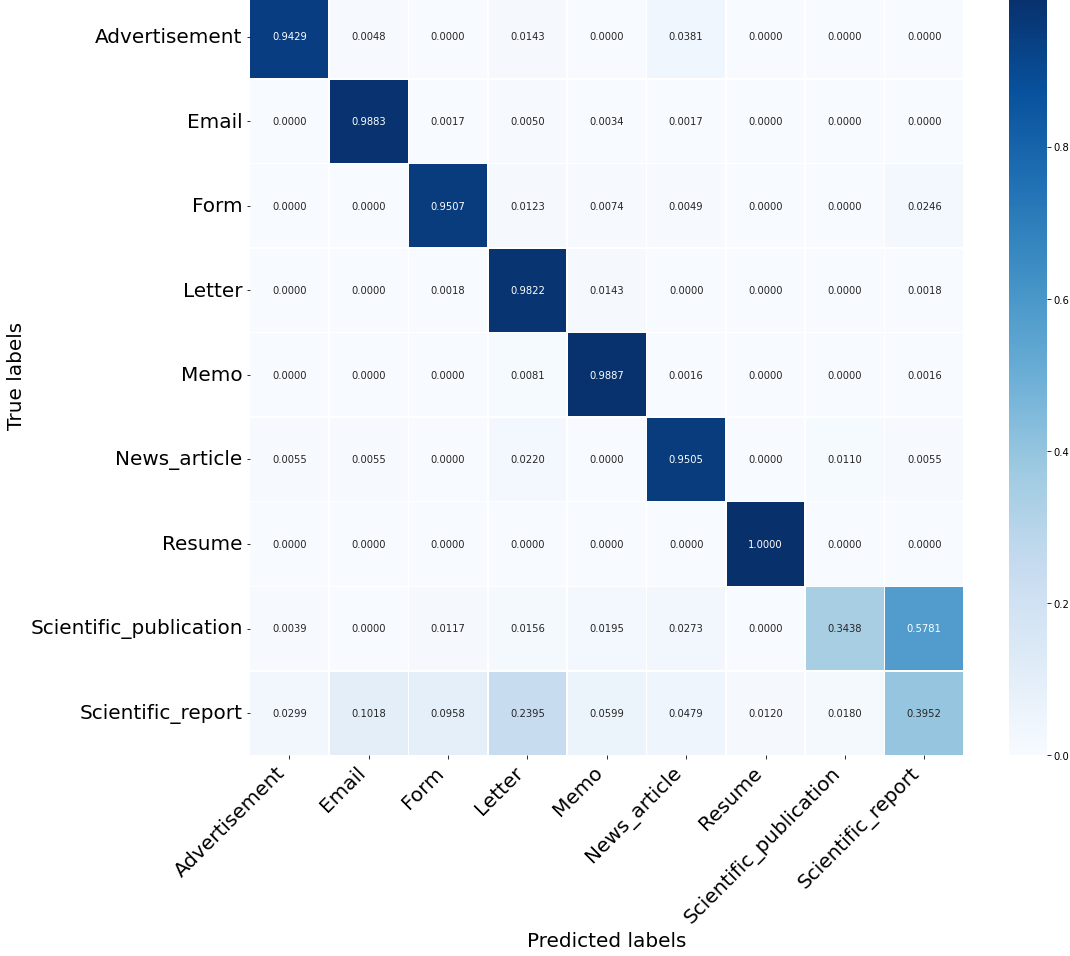}
  \caption{The Image Modality Confusion Matrix of the ML$_{{Tr-KLD}_{Reg}}$) method.}
    \label{fig:CM_EVAL_Image_PML}
\end{figure}

\begin{figure}[ht]
\centering
  \includegraphics[width=1\linewidth]{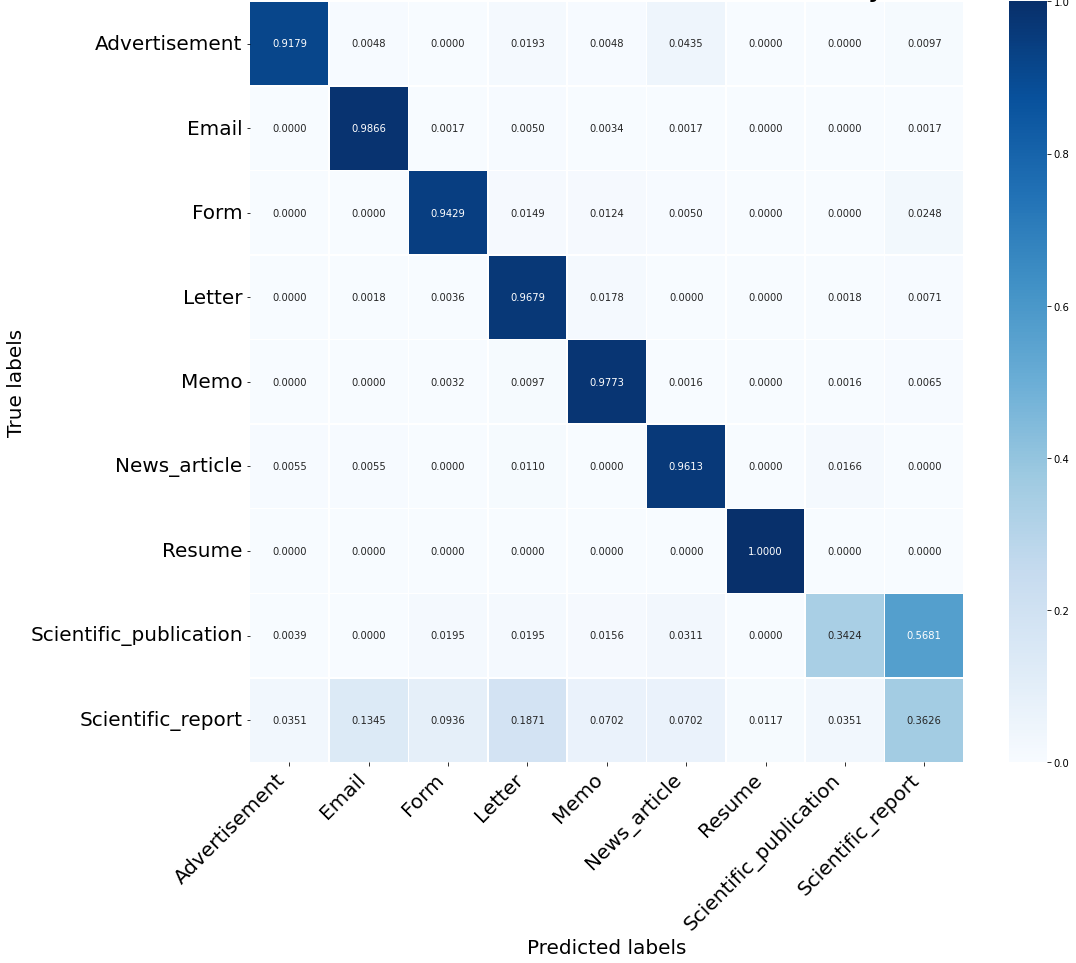}
  \caption{The Text Modality Confusion Matrix of the ML$_{{Tr-KLD}_{Reg}}$ method.}
    \label{fig:CM_EVAL_Text_PML}
\end{figure}

\begin{figure}[ht]
\centering
  \includegraphics[width=1\linewidth]{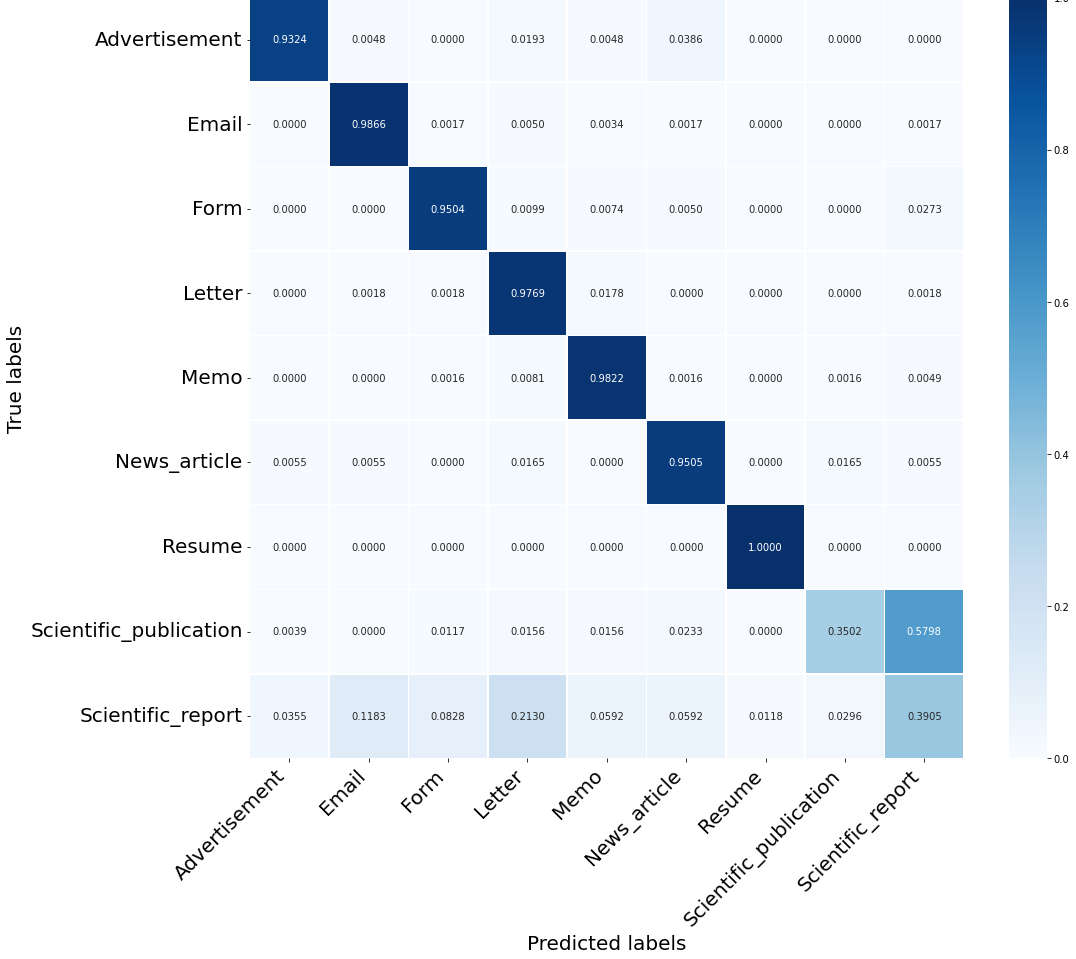}
  \caption{The Multi-Modal Fusion Modality Confusion Matrix of the ML$_{{Tr-KLD}_{Reg}}$ method.}
    \label{fig:CM_EVAL_MultiModal_PML}
\end{figure}

The Confusion matrices in the Figures.~(\ref{fig:CM_EVAL_Image_APML},~\ref{fig:CM_EVAL_Text_APML}, ~\ref{fig:CM_EVAL_MultiModal_APML}), display the generalization ability of our best EAML$_{{Tr-KLD}_{Reg}}$ approach for the image, text, and multi-modal modalities respectively. Symmetrically, the Figures.~(\ref{fig:CM_EVAL_Image_PML},~\ref{fig:CM_EVAL_Text_PML}, ~\ref{fig:CM_EVAL_MultiModal_PML}) refer to the image, text, and multi-modal modalities of the proposed ML$_{{Tr-KLD}_{Reg}}$ method. Note that these methods are trained on RVLCDIP, and evaluated on the Tobacco-3482 dataset.


